\documentclass[10pt,twocolumn,letterpaper]{article}

\usepackage{iccv}              
\usepackage{multirow} 
\usepackage{graphicx}
\usepackage{booktabs}
\usepackage{threeparttable}
\usepackage{makecell}
\usepackage{array,multirow}
\usepackage{setspace}
\usepackage[export]{adjustbox}
\usepackage{pifont}
\usepackage{amsmath}
\usepackage{bm}
\usepackage{booktabs}
\usepackage[accsupp]{axessibility}

\usepackage[pagebackref,breaklinks,colorlinks,allcolors=cvprblue]{hyperref}
\hypersetup{
 colorlinks=true,
 linkcolor=red,
 citecolor=cyan,
 filecolor=magenta, 
 urlcolor=magenta,
 }
\def\paperID{4478} 
\def\confName{ICCV}
\def\confYear{2025}

\title{Beyond Isolated Words: Diffusion Brush for Handwritten Text-Line Generation}

\author{Gang Dai$^{1}$\thanks{Authors contributed equally.}, Yifan Zhang$^{2,3*}$, Yutao Qin$^{1*}$, Qiangya Guo$^{1}$,  Shuangping Huang$^{1,4}$\thanks{Corresponding author}, Shuicheng Yan$^{3}$ \\
$^{1}$South China University of Technology\\
$^{2}$MiroMind AI~
$^{3}$National University of Singapore~
$^{4}$Pazhou Laboratory~
\\
{\tt\small eedaigang@mail.scut.edu.cn}, 
{\tt\small
yifan.zhang@miromind.ai},
{\tt\small eehsp@scut.edu.cn}
}

\begin{document}
\maketitle

\begin{abstract}
Existing handwritten text generation methods primarily focus on isolated words. However, realistic handwritten text demands attention not only to individual words but also to the relationships between them, such as vertical alignment and horizontal spacing. Therefore, generating entire text line emerges as a more promising and comprehensive task. However, this task poses significant challenges, including the accurate modeling of complex style patterns—encompassing both intra- and inter-word relationships—and maintaining content accuracy across numerous characters. To address these challenges, we propose DiffBrush, a novel diffusion-based model for handwritten text-line generation. Unlike existing methods, DiffBrush excels in both style imitation and content accuracy through two key strategies: (1) content-decoupled style learning, which disentangles style from content to better capture intra-word and inter-word style patterns by using column- and row-wise masking;
and (2) multi-scale content learning, which employs line and word discriminators to ensure global coherence and local accuracy of textual content. Extensive experiments show that DiffBrush excels in generating high-quality text-lines, particularly in style reproduction and content preservation. Code is available at \url{https://github.com/dailenson/DiffBrush}
\end{abstract}

\section{Introduction}
\label{sec:intro}

Handwritten text generation aims to automatically synthesize realistic handwritten text images that visually convey a user's personal writing style (\emph{e.g.}, text slant, stroke width, ligatures) while ensuring the content readability. This task has broad applications, including assisting individuals with writing difficulties, accelerating handwritten font design, and enriching data for text recognizer.
Most existing methods~\cite{alonso2019adversarial,fogel2020scrabblegan,bhunia2021handwriting,gan2022higan+,luo2022slogan,pippi2023handwritten,one-dm2024} focus on generating handwritten images at the word level, with few efforts~\cite{DavisMPTWJ20,kang2021content} exploring the generation of complete text lines.  To bridge this gap, our work focuses on high-quality handwritten text-line generation with better control over both style and content.

\begin{figure}[t]   
\vspace{-0.25in} 
\begin{center}
\includegraphics[width=0.99\linewidth]{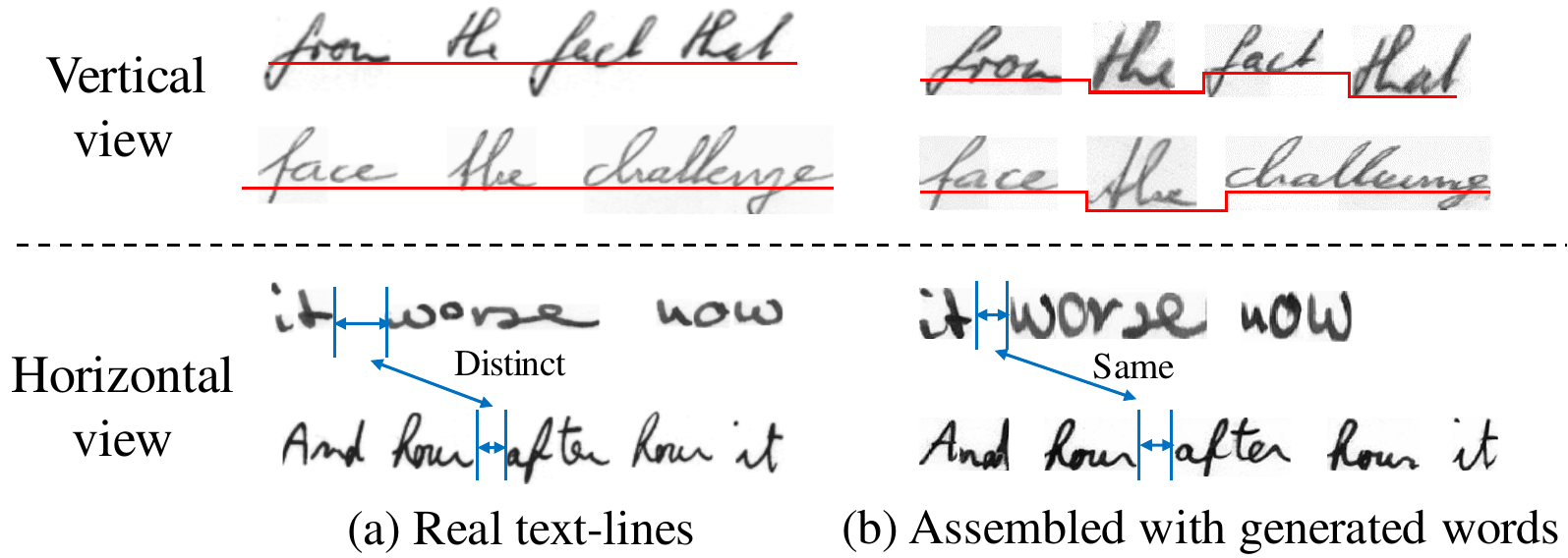}
\end{center}
\vspace{-0.2in} 
\caption{Comparison of handwritten text-lines (a) written by real writers, and (b) assembled with isolated words generated from One-DM~\citep{one-dm2024}. The latter one applies fixed inter-word spacing due to the lack of spacing information in generated words. Red lines indicate the baseline (\emph{i.e.}, the reference line at the bottom of the characters), while blue lines highlight word spacing. 
}
\vspace{-0.25in} 
\label{fig:line_samples}
\end{figure}

Most previous state-of-the-art methods~\cite{gan2022higan+,pippi2023handwritten,nikolaidou2024,one-dm2024} focus on handwritten word generation by using reference images from writers as style inputs and conditioning on character-wise labels or images for content. This allows for the synthesis of handwritten words with controllable styles and specific content. However, as shown in~\cref{fig:line_samples}, generating text at the word level cannot effectively capture the cohesive style of a complete text line: (1) Humans generally maintain consistent vertical alignment across words, while synthesized words often exhibit arbitrary vertical positioning. (2) Different writers have unique word spacing characteristics that are often lost in isolated word generation. 


Direct approaches for text-line generation are relatively limited, with two notable GAN-based methods proposed. TS-GAN~\cite{DavisMPTWJ20} optimizes a global content recognition loss based on the entire generated text-line image, primarily guiding content learning while implicitly influencing style learning. CSA-GAN~\cite{kang2021content}, in contrast, leverages both a content recognition loss and a writer classification loss computed from the generated text-line image, thus better modeling style through writer identity supervision. 
However, both methods suffer from two key limitations: 1) \textbf{Ineffective style extraction}: 
Since both methods jointly optimize content and style from the same model output, the two aspects interfere with each other, preventing effective learning of either. For instance, minimizing content recognition loss inevitably pushes these models to produce easily recognizable outputs with simplified styles (\eg, regular fonts and standard strokes, as shown in~\cref{fig:discussion_ctc}), ultimately hindering the faithful mimicking of diverse handwriting styles. 2) \textbf{Difficulty in maintaining character-level accuracy}:  Ensuring the readability of text lines with numerous characters remains challenging. For instance, in datasets like IAM~\cite{marti2002iam}, where a single text line averages 42 characters—approximately six times the length of a typical word. Optimizing content loss at the text-line level encourages global correctness but may fail to preserve individual character accuracy, making it difficult to maintain content integrity across the entire generated text (cf. $\text{D}_{\text{CER}}$ and $\text{D}_{\text{WER}}$ columns in~\cref{main}).

\begin{figure}[t]  
\vspace{-0.1in} 
\begin{center}
\includegraphics[width=0.95\linewidth]{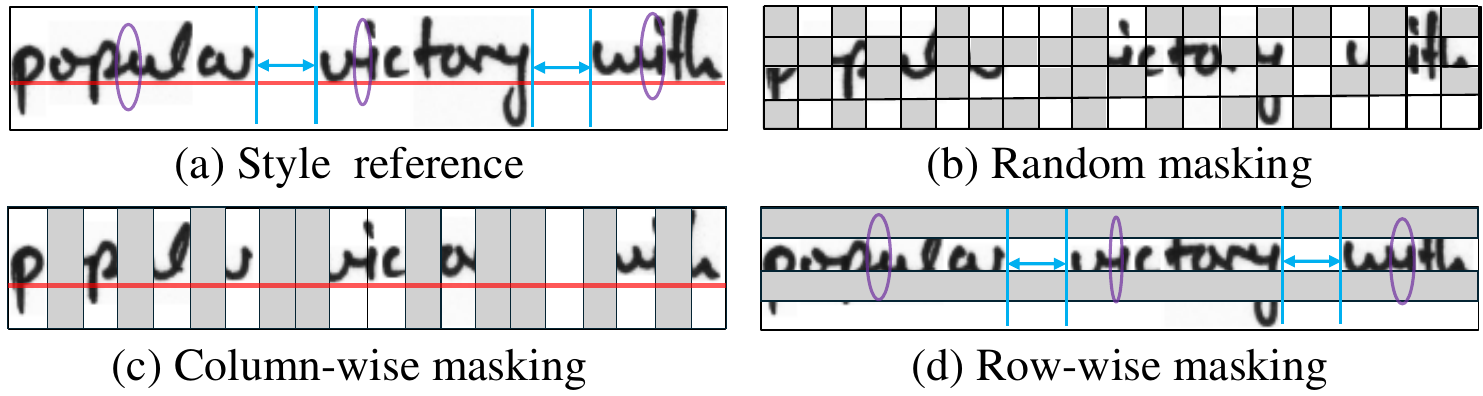}
\end{center}
\vspace{-0.25in} 
\caption{(a) This style reference exhibits rich style patterns, \eg, ligatures, spacing, and vertical alignment alongside undesired content information: ``popular victory with''. (b) Random masking disrupts both style features and content information. (c) Column-wise masking maintains style patterns (\emph{e.g.,} character style and vertical alignment) while removing horizontal content information. (d) Row-wise masking preserves joins and spacing while disrupting vertical content.}
\label{fig:masking}
\vspace{-0.25in}
\end{figure}



To generate handwritten text lines with improved content accuracy and style fidelity, we introduce \textbf{DiffBrush}, a novel diffusion-based approach. Our method incorporates two key strategies: (1) \textbf{content-decoupled style learning}, and (2) \textbf{multi-scale content learning}. 

\begin{figure*}[t]
\vspace{-0.2in}
\begin{center}
\includegraphics[width=0.9\linewidth]{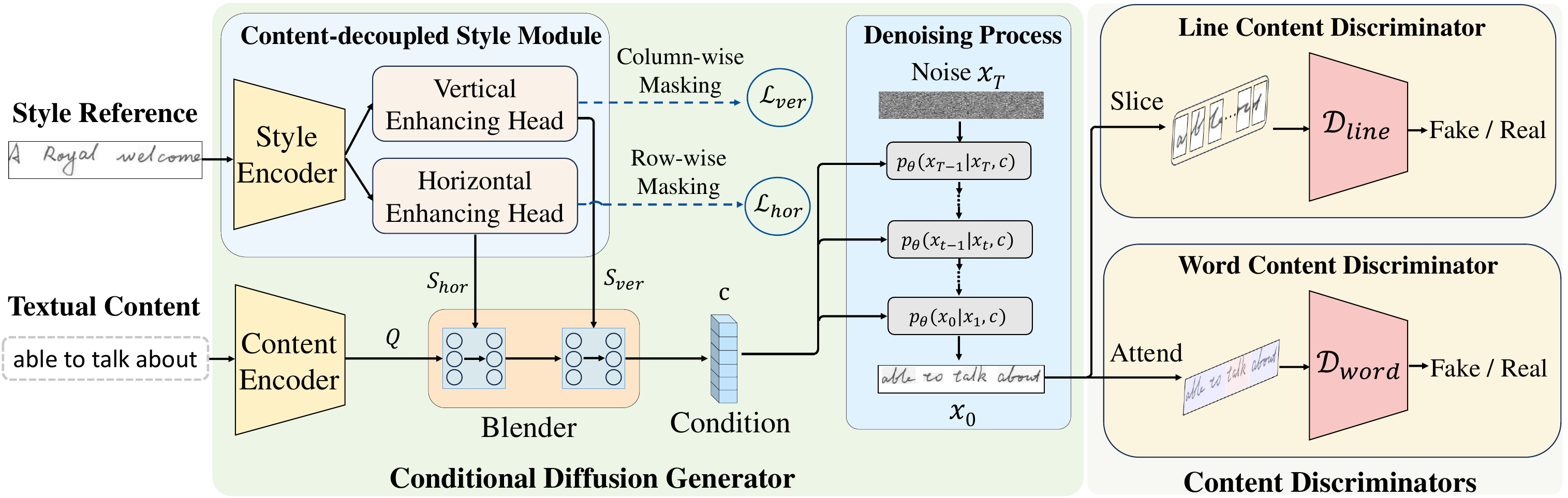}
\end{center}
\vspace{-0.2in}
\caption{Method overview. Our DiffBrush consists of a conditional diffusion generator, a content-decoupled style module, and multi-scale content discriminators. The style module extracts two enhanced style features $S_{hor}$ and $S_{ver}$, which are blended with the content representation $Q$ from the content encoder to construct the condition vector $c$. This condition is then used to guide the denoising process to generate images. To enhance style learning, we explore column- and row-wise masking to effectively eliminate content interference in style modeling. The content discriminators provide feedback at both line and word levels, enhancing the readability of generated text. The slice operation divides the text-line image into horizontal segments, while the attend operation locates word positions in the text-line.}
\vspace{-0.1in}
\label{fig:DiffBrush}
\end{figure*}

Content-decoupled style learning aims to disrupt content information of style references while preserving key style patterns. This eliminates content interference, thus achieving effective one-shot style learning (cf.~\cref{main} and~\cref{fig:main}). A naive approach, such as random masking, fails as it disrupts both content and key style features like word spacing and vertical alignment (cf.~\cref{fig:masking}(b)).
To address this, we propose column- and row-wise masking (cf.~\cref{fig:masking}(c), (d)), which selectively disrupts content while preserving critical style patterns. For style enhancement in vertical direction, as shown in~\cref{fig:DiffBrush}, we apply column-wise masking to the extracted style features, maintaining vertical alignment while removing horizontal content information. A Proxy-NCA loss~\cite{Attias17,kim2020proxy} is then used to enforce style consistency within writers while distinguishing different writers, enabling the vertical enhancing head to refine vertical alignment. For style enhancement in horizontal direction, row-wise masking preserves word and character spacing while disrupting vertical content, allowing the horizontal enhancing head to reinforce horizontal spacing patterns.
This novel masking strategy effectively disentangles style from content, enabling more accurate and independent style representation in handwritten text-line generation.

 
Multi-scale content learning seeks to enhance content accuracy at both global and local levels: at the global level, we preserve character order within a text line to maintain contextual relationships between characters, while at the local level, we ensure the structural correctness of each individual word. To achieve this, we develop a novel multi-scale content discriminator. The line content discriminator segments the text-line image and processes it with a 3D CNN~\cite{tran2015learning} to capture global contextual relationships, encouraging the generator to maintain proper character sequencing. Meanwhile, the word discriminator employs an attention mechanism to isolate individual words and verify their content accuracy, guiding the generator to refine local text content. Our empirical results (cf.~\cref{fig:ablation}) show that this multi-scale content discriminator significantly improves content accuracy without hindering style imitation quality.


Our main contributions include: (1) To the best of our knowledge, DiffBrush is among the first to leverage diffusion generative models for handwritten text-line generation. 
(2) DiffBrush introduces a novel content-decoupled style learning strategy that significantly enhances style imitation, along with a new multi-scale content learning strategy that boosts content accuracy. (3) Extensive experiments on two popular English handwritten datasets (cf.~\cref{main} and~\cref{fig:main}) and one Chinese dataset (cf.~\cref{fig:application}) demonstrate that DiffBrush significantly outperforms state-of-the-arts. 

\section{Related Work} 
\label{gen_inst}

Handwritten text generation methods are generally divided into online and offline: the former synthesizes dynamic stroke sequences, while the latter generates static text images. With the advancement of deep learning, Transformer decoders~\cite{dai2023disentangling} and diffusion models~\cite{chirodiff,luhman2020diffusion,ren2023diff} have been used for synthesizing online handwritten text. However, as highlighted in recent studies~\cite{bhunia2021handwriting,pippi2023handwritten,one-dm2024}, online methods require temporal data (\eg, coordinate points and writing orders) collected from a digital stylus pen and cannot synthesize stroke width, ink color like offline methods. In light of this, this paper focuses on offline handwriting generation.

The advent of Generative Adversarial Networks~\cite{liu2021deep,huang2022agtgan} has accelerated the development of offline handwritten text generation. Early works~\cite{alonso2019adversarial,fogel2020scrabblegan} use character labels as content inputs and random noise as style inputs to synthesize handwritten words with controllable content and random styles. To enhance style control, SLOGAN~\cite{luo2022slogan} conditions style inputs on fixed writer IDs but fails to mimic unseen styles. Unlike them, GANwriting~\cite{kang2020ganwriting} and HWT~\cite{bhunia2021handwriting} employ CNN or transformer encoder to extract style features from style references and are thus capable of imitating any styles.  
Further, VATr~\cite{pippi2023handwritten} utilizes symbol images as content representations, enabling character generation beyond the training charset.
In contrast to the above word-focused methods, TS-GAN~\cite{DavisMPTWJ20} and CSA-GAN~\cite{kang2021content} are developed to synthesize handwritten text-lines. However, they struggle to produce satisfactory results due to design drawbacks in style learning and content supervision.


The rapid development of diffusion models~\cite{dhariwal2021diffusion,WangZHCZ23,xu2024ufogen,liang2024diffusion,hu2025replaycad} offers new potential for handwritten text generation. However, some early attempts~\cite{zhu2023conditional,2023wordstylist}, which condition denoising process on fixed writer labels, cannot mimic unseen handwriting styles. To address this, DiffusionPen~\cite{nikolaidou2024} and One-DM~\cite{one-dm2024} extract style information from reference images, and then merge this information with the textual content to guide the denoising process. However, regarding text content readability, One-DM simply incorporates a text recognizer with a CTC decoder, while DiffusionPen neglects this challenge entirely. Different from them, we propose a novel multi-scale content learning strategy, significantly enhancing text content readability. Moreover, previous diffusion methods~\cite{zhu2023conditional,2023wordstylist,nikolaidou2024,one-dm2024} focus on generating isolated words whereas our DiffBrush aims for entire text-line generation. We discuss more related works about diffusion methods for general image generation in Appendix~\ref{sec:diffusion}.


\section{Problem Statement and Preliminaries}

\textbf{Problem statement.}
Given a text string $\mathcal{A}$ and a style reference $s_i$ randomly sampled from an exemplar writer $w_i \in \mathcal{W}$, we aim to synthesize a handwritten text-line image $x$ that captures the unique calligraphic style of $w_i$ while accurately preserving the content of $\mathcal{A}$. Here, $\mathcal{A} = \{a_i\}_{i=1}^{L}$ represents a sequence of length $L$, where each $a_i$ is a Unicode character, including lowercase and uppercase letters, digits, punctuation. The key challenges lie in accurately capturing handwriting styles, including both intra- and inter-word patterns from the style reference, while ensuring the readability of text-lines that typically contain numerous characters.

\vspace{0.1in}
\noindent\textbf{Conditional diffusion model.} The diffusion model~\cite{ho2020denoising} generates realistic images by progressively denoising a random Gaussian noise input. To achieve {controllable generation}, the {conditional diffusion model}~\cite{chen2025human,rombach2022high,zhang2023hipa,zhang2025matrix,zheng2024memo,zhang2023expanding} incorporates a condition signal $c$ to guide the denoising process. Starting from pure Gaussian noise $x_T \sim \mathcal{N}(0, \mathcal{I})$, a denoising network $p_{\theta}$ iteratively refines the image over multiple timesteps to produce the target image $x_0$. The network $p_{\theta}$, typically based on a {U-Net architecture}~\cite{RonnebergerFB15}, integrates $c$ via cross-attention or adaptive modulation layers. The training objective minimizes the mean squared error (MSE) between the predicted and real images:
$\mathcal{L}_{\text{diff}} = \mathbb{E}_{x_t, c} \left[ \| x_{0} - x_{\text{real}} \|^2 \right].$ 
By leveraging condition signals such as text prompts and reference images, conditional diffusion models enable fine-grained control over the generation.

\section{Method}

\begin{figure*}[t]
\vspace{-0.2in}
\begin{center}
\includegraphics[width=0.85\linewidth]{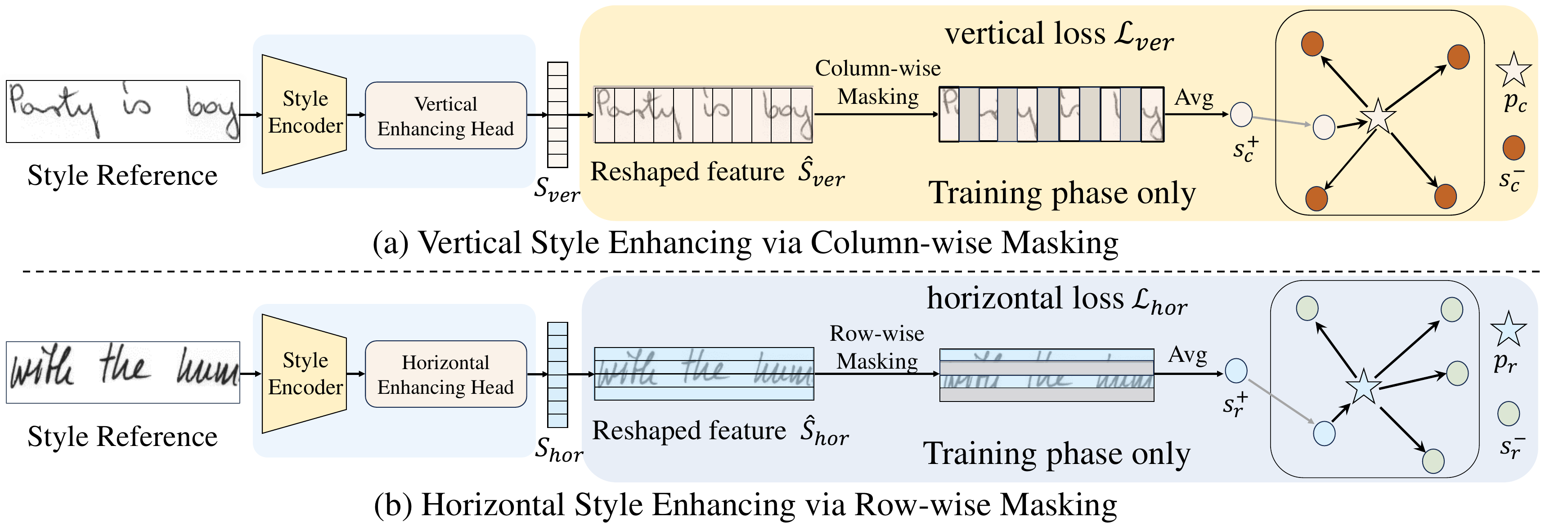}
\vspace{-0.2in}
\end{center}
\caption{Style learning via column- and row-wise masking. (a) An improved vertical style feature $S_{ver}$ is extracted by the vertical enhancing head, guided by a vertical loss $\mathcal{L}_{ver}$. Specifically, during training, we begin reshaping $S_{ver}$ into spatial feature $\hat{S}_{ver}$ and perform column-wise masking on $\hat{S}_{ver}$. After average pooling, $s_{c}^+$ is drawn closer to its corresponding writer proxy $p_{c}$, while the negatives $s_{c}^-$ belonging to different writers are pushed away by $p_{c}$. (b) Similarly, a style feature $S_{hor}$ is extracted by the horizontal enhancing head. We reshape $S_{hor}$ and conduct row-wise masking. After pooling, $s_{r}^+$ is linked to its writer proxy $p_{r}$, and negatives $s_{r}^-$ are pushed away.} 
\vspace{-0.1in}
 
\label{fig:sample}

\end{figure*}

 \subsection{{Overall Scheme}} \label{sec:overall}

To generate handwritten text-lines with enhanced content accuracy and style fidelity, we propose DiffBrush, a novel conditional diffusion generation method. As shown in~\cref{fig:DiffBrush},  the architecture of DiffBrush consists of three main components: content-decoupled style module, conditional diffusion generator, and multi-scale content discriminators. 

The content-decoupled style module $\xi_{style}$ aims to better capture the text-line styles of exemplar writers. To achieve this, we introduce a content-decoupled style learning strategy (cf.~\cref{sec:style}), which leverages two novel content-masking techniques and a style learning loss $\mathcal{L}_{style}$ to enhance text-line style modeling (cf.~\cref{fig:DiffBrush}). The extracted style features are then fused with content features from a content encoder to form the condition vector $c$ within a blender module. Both the content encoder and blender module are designed based on One-DM~\cite{one-dm2024}, with further extensions (cf. Appendix ~\ref{imp} for details). Guided by $c$, the conditional diffusion generator $\mathcal{G}$ performs denoising process to synthesize realistic handwritten text-line image $x_{0}$. 

However, training the generator $\mathcal{G}$ with solely the diffusion loss $\mathcal{L}_{diff}$ is insufficient to ensure the content readability of generated text lines. To address this, we introduce a multi-scale content learning strategy  (cf.~\cref{sec:content}). Specifically, we develop a multi-scale discriminator $\mathcal{D}$ to evaluate the content correctness at both the line and word levels, thus providing more fine-grained content supervision $\mathcal{L}_{conent}$ for content adversarial learning between $\mathcal{G}$ and $\mathcal{D}$.
 
To summarize,  the overall training objectives of our DiffBrush combines all three loss functions:
\begin{equation}
\begin{aligned}
\mathcal{L}_{\mathcal{G}}= \mathcal{L}_{diff} + \mathcal{L}_{style} +  \lambda \mathcal{L}_{content},
\end{aligned}
\label{eq1}
\end{equation}
where $\lambda$ serves as a trade-off factor, and we empirically set it to $0.05$ in training.

\begin{figure*}[t]
\vspace{-0.1in}
\begin{center}
\includegraphics[width=0.85\linewidth]{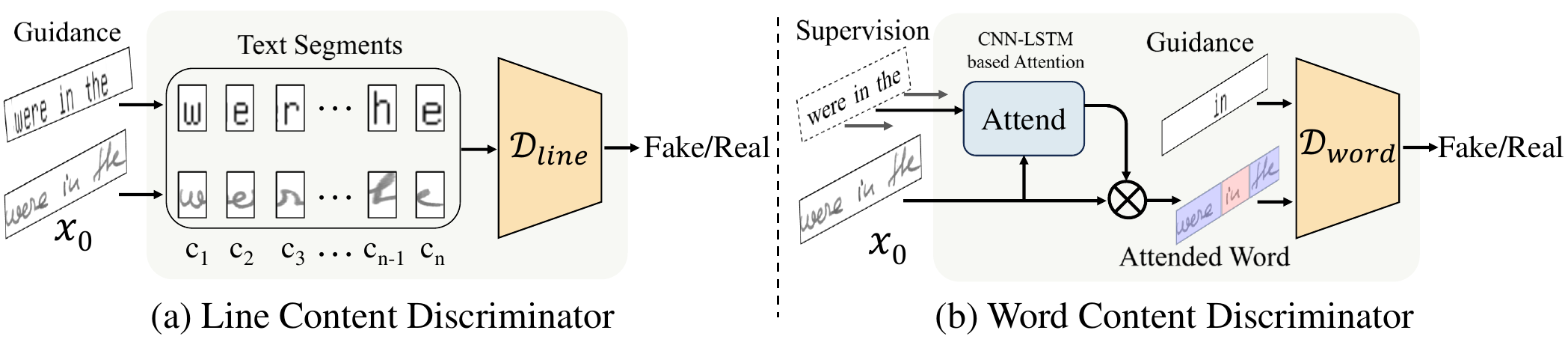}
\end{center}
\vspace{-0.2in}
\caption{Illustration of the multi-scale content discriminators. (a) For the line content discriminator $\mathcal{D}_{line}$, we concatenate the generated text-line image $x_0$ and guidance image along the channel dimension, slice the result into $n$ segments, and then process them with a 3D CNN~\cite{tran2015learning} to integrate context information. Finally, the line discriminator $\mathcal{D}_{line}$ evaluates each segment as real or fake. (b) For the word content discriminator $\mathcal{D}_{word}$, we utilize an attention module~\cite{shi2018aster} to attend to individual words within the generated text line image $x_{0}$, and then input these words, along with corresponding content guidance, into $D_{word}$ for realism discrimination.}
\vspace{-0.1in}
\label{fig:Discriminators}
\end{figure*}

\subsection{Content-decoupled Style Learning}\label{sec:style} 

As discussed in Section~\ref{sec:intro}, text-line style learning is hindered by content interference, leading to ineffective style extraction. To address this, we propose to disrupt content information (by masking) to eliminate content interference  during style learning, forcing the model to focus on  essential style patterns. Specifically, as shown in~\cref{fig:DiffBrush}, style samples are first processed by a CNN-Transformer style encoder to obtain an initial style feature $S$ with rich calligraphic attributes. To refine style representation, we introduce two dedicated style-enhancing heads, each incorporating a standard self-attention layer to extract fine-grained style features: $S_{ver}$ (vertical) and $S_{hor}$ (horizontal). These features are learned by using row- and column-wise masking for content disruption. The overall content-decoupled style loss $\mathcal{L}_{style}$ is formulated as the sum of a vertical enhancing loss $\mathcal{L}_{ver}$ and a horizontal enhancing loss $\mathcal{L}_{hor}$: 
\begin{equation}
\begin{aligned}
\mathcal{L}_{style}=\mathcal{L}_{ver} + \mathcal{L}_{hor}.
\end{aligned}
\end{equation}
 
\noindent\textbf{Vertical style enhancing via column-wise masking.} The vertical enhancing head aims to enhance style learning in the vertical direction (\eg, vertical alignment patterns) via column-wise masking. Specifically, as shown in~\cref{fig:sample}(a), we perform masking on $S_{ver}$ by first reshaping the sequential feature $S_{hor}$ back into spatial feature $\hat{S}_{ver} \in \mathbb{R}^{h \times w \times c}$. We then divide $\hat{S}_{ver}$ into several columns and randomly mask a subset of columns with equal probability to obtain an average feature $s_{c} \in \mathbb{R}^{h \times n \times c}$, where $n=w \cdot \rho$ and $\rho$ is the masking ratio. After column-wise masking, we adopt a Proxy-NCA loss~\cite{Attias17,kim2020proxy} for style learning, which enforces style consistency within writers while distinguishing different writers. Specifically, our vertical enhancing loss $\mathcal{L}_{ver}$ assigns a proxy to each writer, treating it as an anchor to cluster the masked style features of the same writer while  pushing apart those of different writers:
\begin{align}
\resizebox{0.9\linewidth}{!}{%
    $\begin{aligned}
    \mathcal{L}_{ver} = & \frac{1}{|P_{c}^+|} \sum_{p_{c} \in P_{c}^+} \log \left( 1 + \sum_{s_{c} \in S_{c}^+} e^{- f_{c}^{+}} \right)
    + & \frac{1}{|P_{c}|} \sum_{p_{c} \in P_{c}} \log \left( 1 + \sum_{s_{c}  \in S_{c}^-} e^{f_{c}^{-}} \right),
    \end{aligned}$
}
\end{align}
where $S_{c}=\{s_{c}^{i}\}_{i=1}^{N}$ is a batch of masked style features, $P_{c}$ denotes the set of proxies of all writers, and $P_{c}^{+}$ refers to the set of writers present in the current batch. For each proxy $p_{c}$, $S_{c}$ is divided into a positive set $S_{c}^{+}$, consisting of $s_{c}$ from the same writer as $p_{c}$, and a negative set $S_{c}^{-} = S_{c} - S_{c}^{+}$. The similarity between positive pairs is $f_{c}^{+} = {\alpha (g(s_{c}, p_{c}) - \delta)}$ for $s_{c} \in S_{c}^+$, and that of negative pairs is $f_{c}^{-} = {\alpha (g(s_{c}, p_{c}) + \delta)}$ for $s_{c} \in S_{c}^-$, where  $g({\cdot})$ is the cosine similarity, $\delta > 0$ is a margin and $\alpha$ is a scaling factor.


\vspace{0.05in}
\noindent\textbf{Horizontal style enhancing via row-wise masking.} 
The horizontal enhancing head aims to enhance style learning in the horizontal direction (\eg, word and character spacing) via row-wise masking. As shown in~\cref{fig:sample}(b), the masking operation is conducted in a similar way as column-wise making. 
Specifically, we reshape the sequential feature $S_{hor}$ back into a spatial feature $\hat{S}_{hor} \in \mathbb{R}^{h \times w \times c}$ and then conduct random row-wise masking to obtain $s_{r} \in \mathbb{R}^{m \times w \times c}$, where $m = h \cdot \rho$. Our horizontal enhancing loss $\mathcal{L}_{hor}$ pulls the masked style features of the same writer  together while pushing those of different writers apart:
\begin{align}
\resizebox{0.9\linewidth}{!}{%
    $\begin{aligned}
    \mathcal{L}_{hor} = \frac{1}{|P_{r}^+|} \sum_{p_{r} \in P_{r}^+} \log \left( 1 + \sum_{s_{r} \in S_{r}^+} e^{-f_{r}^{+}} \right)
    + & \frac{1}{|P_{r}|} \sum_{p_{r} \in P_{r}} \log \left( 1 + \sum_{s_{r} \in S_{r}^-} e^{f_{r}^{-}} \right),
    \end{aligned}$%
}
\end{align}
where we assign a proxy $p_{r}$ to each writer and link it with all masked results $S_{r}$. The similarity between positive pairs is $f_{r}^{+} = {\alpha (g(s_{r}, p_{r}) - \delta)}$ for $s_{r} \in S_{r}^+$, and that of negative pairs is $f_{r}^{-} = {\alpha (g(s_{r}, p_{r}) + \delta)}$  for $s_{r} \in S_{r}^-$.


\subsection{Multi-Scale Content Learning}\label{sec:content}

Existing methods~\cite{DavisMPTWJ20,kang2021content,gan2022higan+,pippi2023handwritten,one-dm2024} rely on content recognition losses to enhance the content readability of generated handwriting images. However, these losses, applied at the text-line level, prioritize global correctness but often fail to ensure character-level accuracy. For instance, in datasets like IAM~\cite{marti2002iam}, where a single text line averages 42 characters—approximately six times the length of a typical word, maintaining content integrity across the entire generated text becomes challenging. 
To address this problem, we introduce a multi-scale content learning strategy that provides finer-grained content supervision at both global (line) and local (word) levels for content adversarial training. As shown in~\cref{fig:Discriminators}, the line content discriminator $\mathcal{D}_{line}$ evaluates the overall character order with a line discrimination loss $\mathcal{L}_{line}$, while the word content discriminator $\mathcal{D}_{word}$ ensures character-level accuracy through a word discrimination loss $\mathcal{L}_{word}$.  The overall multi-scale content loss is then formulated as:
\begin{equation}
\begin{aligned}
\mathcal{L}_{content}=\mathcal{L}_{line}+\mathcal{L}_{word}.
\end{aligned}
\end{equation}

\vspace{0.05in}
\noindent\textbf{Line content discriminator.} As shown in~\cref{fig:Discriminators}(a), given the generated image $x_0$ after diffusion and the content guidance $I_{line}$ without style information, the line discriminator $\mathcal{D}_{line}$ aims to determine whether the overall character order in $x_0$ matches that in $I_{line}$. Firstly, we set the wider of the $I_{line}$ and $x_0$ as the benchmark, padding the narrower one with white background pixels for width alignment. We then concatenate $x_0$ and $I_{line}$ along the channel dimension, and then slice the concatenated result into $n$ non-overlapping segments $\{c_i\}_{i=1}^n$ from left to right. Afterwards, a 3D CNN~\cite{tran2015learning} based discriminator $\mathcal{D}_{line}$ processes $\{c_i\}_{i=1}^n$ to incorporate global context information of characters, and determines whether this output is real or fake, providing feedback for the overall character order. The line discriminator loss $\mathcal{L}_{line}$ is formulated as: 
\begin{equation}
\resizebox{0.9\linewidth}{!}{%
    $\begin{aligned}
    \mathcal{L}_{line} = \log(\mathcal{D}_{line}(I_{line}, x_{real})) + \log(1 - \mathcal{D}_{line}(I_{line}, x_0)).
    \end{aligned}$%
}
\end{equation}

\vspace{0.05in}
\noindent\textbf{Word content discriminator.} Compared to the line discriminator $\mathcal{D}_{line}$, the word discriminator $\mathcal{D}_{word}$ is designed to ensure that the text structure is correctly generated at the word level. However, accurately locating word positions within a whole text-line $x_0$ is non-trivial. Motivated by ASTER~\cite{shi2018aster}, we utilize an attention module with a CNN-LSTM architecture to obtain word positions.

As shown in~\cref{fig:Discriminators}(b), given the generated image $x_0$ after diffusion, a CNN encoder first extracts spatial features $F_{map} \in \mathbb{R}^{h \times w \times c}$ from $x_0$, which is flattened into sequential features $H \in \mathbb{R}^{l \times c}$, where $l=h \times w$. The LSTM decoder then takes $x_0$ and a start-of-sequence (SOS) token as input, sequentially outputting attention maps for character positions until the end-of-sequence (EOS) token is reached. The character-level attention maps are then concatenated into word-level attention maps $A=\{a_t\}_{t=1}^T$ (cf. Figure~\ref{fig:attention_heatmap} in Appendix), where $a_t \in \mathbb{R}^{h \times w}$  and $T$ denotes the number of words in the text-line. Based on the attention maps, we extract attended words $\{x_{word}^t\}_{t=1}^T$, with $x_{word}^t=a_t \cdot x_{0}$. 
Lastly, each $x_{word}$ and its corresponding content guidance $I_{word}$ are fed into $\mathcal{D}_{word}$ for discrimination, which provides word-level content feedback for the generator to refine the local content readability. Specifically, the word discriminator loss $\mathcal{L}_{word}$ is formulated as: 
\begin{align}
\resizebox{1\linewidth}{!}{%
    $\begin{aligned}
    \mathcal{L}_{word} = &
    \sum_{i=1}^{T} \log(\mathcal{D}_{word}(I_{word}^i, x_{real}^i)) + \sum_{i=1}^{T} \log(1 - \mathcal{D}_{word}(I_{word}^i, x_{word}^i)),
    \end{aligned}$%
}
\end{align}
where $i$ represents the $i$-th word in a text line.

\begin{table*}[t]
\vspace{-0.2in}
\centering
\scalebox{0.8}{
\begin{tabular}{lcccccccc}
\toprule
Datasets & Method & Shot &  HWD 
$\downarrow$ & $\text{D}_{\text{CER}}$ $\downarrow$ & $\text{D}_{\text{WER}}$ $\downarrow$ & FID $\downarrow$ & IS $\uparrow$ & GS $\downarrow$ \\
\midrule
\multirow{5}{*}{IAM} & TS-GAN~\cite{DavisMPTWJ20}          & one & 2.11  & 44.20 & 87.13 & 16.76 & 1.76 & 2.87 $\times 10^{-2}$ \\
& CSA-GAN~\cite{kang2021content}         & few & 2.25  & 42.27 & 84.14 & 13.52 & 1.74 & 1.62 $\times 10^{-2}$ \\
& VATr~\cite{pippi2023handwritten}            & few & 1.87  & 28.80 & 71.77 & 12.51 & 1.69 & 1.45 $\times 10^{-2}$ \\
&DiffusionPen~\cite{nikolaidou2024} & few & 1.72  & 54.75 & 84.70 & 10.24 & 1.83 & 6.42 $\times 10^{-3}$ \\
& One-DM~\cite{one-dm2024}          & one & 1.80  & 20.91 & 54.27 & 10.60 & 1.82 & 8.42 $\times 10^{-3}$ \\
& Ours            & one & \textbf{1.41}  & \textbf{8.59}  & \textbf{28.60} & \textbf{8.69}  & \textbf{1.85} & \textbf{2.35} \bm{$\times 10^{-3}$} \\

\midrule
\multirow{4}{*}{CVL} 
& CSA-GAN~\cite{kang2021content}         & few & 1.72  & 41.64 & 72.02 & 8.71 & 1.48 & 6.71 $\times 10^{-2}$ \\
& VATr~\cite{pippi2023handwritten}            & few & 1.50  & 38.49 & 66.33 & 9.04 & 1.44 & 1.43 $\times 10^{-1}$ \\
&DiffusionPen~\cite{nikolaidou2024} & few & 1.32 & 55.94 & 88.36 & 11.90 & 1.59  & 5.08 $\times 10^{-2}$ \\
& One-DM~\cite{one-dm2024}          & one & 1.47  & 32.42 & 63.35 & 11.95 & 1.46 & 1.29 $\times 10^{-1}$ \\
& Ours            & one & \textbf{1.06}  & \textbf{20.92}  & \textbf{36.38} & \textbf{7.57}  & \textbf{1.70} & \textbf{2.96} \bm{$\times 10^{-2}$} \\  

\bottomrule
\end{tabular}
} 
\caption{Comparisons with baselines on handwritten text-line generation on IAM and CVL. All methods are trained on the same training set and evaluated using the same protocols. The ``Shot'' column indicates the number of text-line style references required for each method.}
\label{main}
\vspace{-0.1in}
\end{table*}

\section{Experiments}\label{others}
\subsection{Experimental Settings}
\textbf{Evaluation dataset.} To evaluate our DiffBrush in generating handwritten text-line, we use the widely adopted handwriting datasets IAM~\cite{marti2002iam} and CVL~\cite{kleber2013cvl}. IAM contains 13,353 English text-line images belonging to 657 unique writers. Following the protocol of CSA-GAN~\cite{kang2021content}, we use text-lines from 496 writers for training and the remaining 161 writers for testing. CVL dataset consists of handwritten text-lines from 310 writers in both English and German. For our experiments, we use the English portion, consisting of 11,007 text-lines, and follow the standard CVL split, with 283 writers for training and 27 for testing. In all experiments, we resize the images to a height of 64 pixels while preserving their aspect ratio, as done in previous works\cite{DavisMPTWJ20,kang2021content,one-dm2024}. To manage varying widths, images with a width smaller than 1024 pixels are padded, whereas those exceeding 1024 are resized to a fixed size of 64 $\times$ 1024. We also evaluate DiffBrush on popular Chinese dataset CASIA-HWDB (2.0–2.2)~\cite{liu2011casia} (cf. Appendix~\ref{Chinese} for more details.)

\begin{figure*}[t]
\begin{center}
\vspace{-0.1in}
\includegraphics[width=0.9\linewidth]{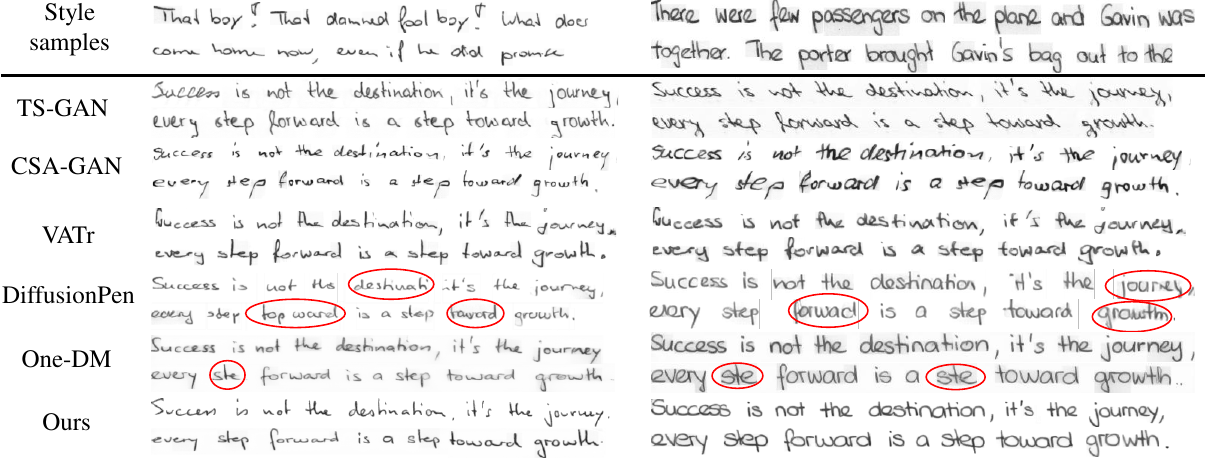}
\end{center}
\vspace{-0.2in}
\caption{Qualitative comparisons between our method and state-of-the-art approaches for handwritten text-line generation, conditioned on out-of-vocabulary (OOV) textual content and unseen styles from the IAM test dataset. We use the same guiding text, ``Success is not the destination, it's the journey, every step forward is a step toward growth.'' for all methods, instructing them to generate the text in different handwriting styles. The red circles highlight missing characters or structural errors. Better zoom in 200$\%$.} 
\label{fig:main}
\end{figure*}

\noindent\textbf{Evaluation metrics.} 1) We use the newly proposed Handwriting Distance (HWD)~\cite{PippiQCC23}, specifically designed for handwriting style evaluation. HWD computes the Euclidean distance between features extracted by a VGG16 network pre-trained on a large corpus of handwritten text images. 2) For content evaluation, we follow latest studies~\cite{rethinking,nikolaidou2024,one-dm2024} by using the generated training sets from each method to train an OCR system~\cite{retsinas2022best} and report its recognition performance on the real test set in terms of CER and WER. We name these new evaluation metrics $\text{D}_{\text{CER}}$ and $\text{D}_{\text{WER}}$, with further discussions provided in Appendix~\ref{content}. 
3) We use Fréchet Inception Distance (FID)~\cite{heusel2017gans}, Inception Score (IS)~\cite{salimans2016i}, and Geometry Score (GS)~\cite{2018geometry} to measure the visual quality of generated images. 4) We also conduct user studies to quantify the subjective quality of the generated handwritten text-line images in Appendix~\ref{user_study}.

\noindent \textbf{Implementation details.} In all experiments, we use a randomly selected text-line sample as the style reference. In DiffBrush, both the style and content encoders are based on a ResNet18, followed by 2 transformer encoder layers. 
Line discriminator uses three 3D  convolution layers, and word discriminator has three 2D convolution layers. The model is trained for 800 epochs on eight RTX 4090 GPUs using the AdamW optimizer with a learning rate of $10^{-4}$. We set masking ratio $\rho$ to 0.5 after a grid search (cf.~\cref{tab:masking} in Appendix). We randomly drop the condition $c$ with the probability $0.1$ for classifier-free training~\cite{ho2022classifier}. During inference, we adopt a classifier-free guidance scale of 0.2 and use DDIM~\cite{SongME21} with 50 steps to accelerate the process. More details are put in Appendix~\ref{imp}.

\noindent \textbf{Compared Methods.}
We compare DiffBrush with state-of-the-art handwritten text-line generation methods, including TS-GAN~\cite{DavisMPTWJ20}, CSA-GAN~\cite{kang2021content}, and advanced word-level handwritten text generation approaches like VATr~\cite{pippi2023handwritten}, DiffusionPen~\cite{nikolaidou2024} and One-DM~\cite{one-dm2024}. 
We provide implementation details of word-level methods in \cref{non-text}.

\subsection{Main Results}

\begin{figure*}[t]
\begin{center}
\includegraphics[width=0.62\linewidth]{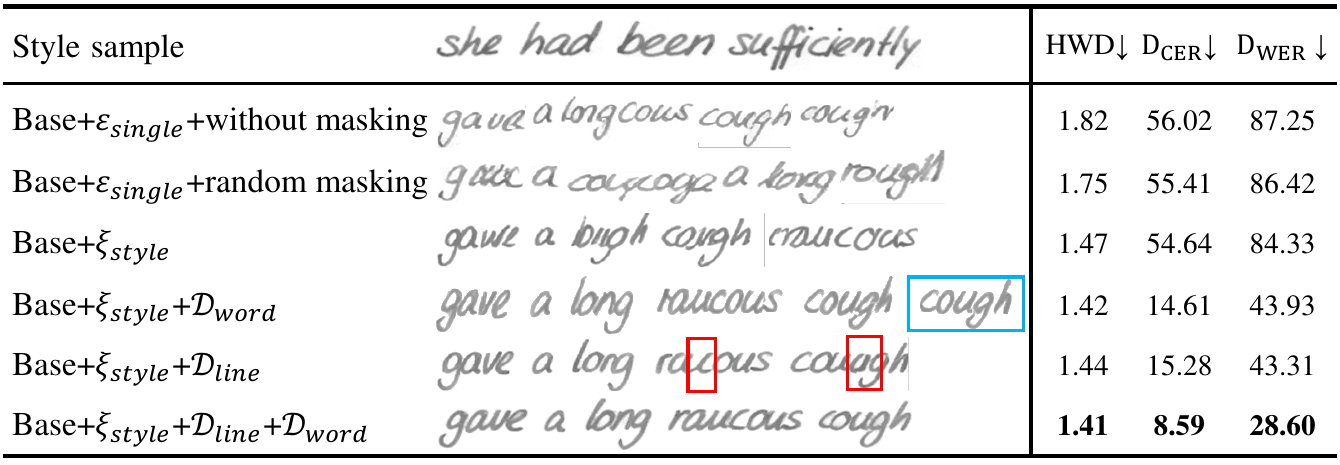}
\end{center}
\vspace{-0.2in}
\caption{Ablation studies of the content-decoupled style module ($\xi_{style}$) and the multi-scale content discriminators (\emph{i.e.,} $\mathcal{D}_{line}$ and $\mathcal{D}_{word}$) based on IAM test set. $\varepsilon_{single}$ denotes a single CNN-Transformer style encoder (ResNet18 followed by 3 transformer encoder layers). The red boxes highlight failure instances of structure preservation, whereas the blue box points out an incorrect repetitive word.}
\label{fig:ablation}
\vspace{-0.1in}
\end{figure*}

\textbf{Styled handwritten text-line generation.} We assess DiffBrush for generating handwritten text-line images with desired style and specific content. To quantify style similarity, following CSA-GAN~\cite{kang2021content}, we generate text-line images for each method using style information from test set and content input from a subset of WikiText-103~\cite{merity2016pointer}. We then calculate the HWD between the generated and real samples for each writer, and finally average the results.

The quantitative results in~\cref{main} show that DiffBrush outperforms all state-of-the-art methods on both IAM and CVL datasets. Specifically, it improves HWD by $18.02\%$ ($1.72 \rightarrow 1.41$) on IAM and $19.69\%$ ($1.32 \rightarrow 1.06$) on CVL compared to the second-best method, highlighting its superior style imitation ability. Moreover, DiffBrush achieves significantly lower $\text{D}_{\text{CER}}$ and $\text{D}_{\text{WER}}$ on both IAM and CVL datasets, demonstrating its advantage in content readability. {In contrast, DiffusionPen~\cite{nikolaidou2024} yields the highest $\text{D}_{\text{CER}}$ and $\text{D}_{\text{WER}}$ due to ineffective content supervision.}

We provide qualitative results to intuitively explain the benefit of our DiffBrush in~\cref{fig:main}. TS-GAN struggles to accurately capture the style patterns of reference samples, like ink color and stroke width. CSA-GAN produces samples that lack style consistency, including inconsistent character slant, ink color, and stroke width. VATr has difficulty maintaining vertical alignment between words in the synthesized text lines. {DiffusionPen struggles to ensure the content readability of generation results.} One-DM occasionally generates text lines with missing or incorrect characters. Conversely, our DiffBrush excels at generating precise character details while maintaining overall consistency.

\vspace{0.05in}
\noindent \textbf{Style-agnostic text-line generation.}
We further evaluate DiffBrush’s ability to generate realistic handwritten text-line images, independent of style imitation. Following TS-GAN~\cite{DavisMPTWJ20}, each method generates 25k random text-line images to calculate FID against all training samples, and 5k random samples for GS calculation, compared with 5k samples from the test set. Besides, we generate the entire test set using each method and evaluate the results using the IS. As shown in~\cref{main}, DiffBrush achieves the highest performance across FID, IS, and GS metrics on both IAM and CVL datasets, further demonstrating its ability to generate superior-quality handwritten text-line images.

\subsection{Analysis}
In this section, we conduct ablation studies to analyze our DiffBrush. We provide more analyses in Appendix, including generalization evaluation on various
style backgrounds, \textbf{failure case analysis},
ablation results on masking ratio, enriching datasets to train recognizer, style interpolation results, style evaluation results in terms of WIER~\cite{gan2022higan+,zdenek2023handwritten}, discussions about fine-grained style learning.

\vspace{0.05in}
\noindent \textbf{Quantitative evaluation of style module and content discriminators.}
We perform multiple ablation studies
to analyze different components. Quantitative
results in~\cref{fig:ablation} reveal that: (1) Compared to two variants—a basic style encoder without masking, and with random masking (best masking ratio 0.5)—our style module $\xi_{style}$ significantly improves HWD by $19.23\%$ ($1.82\rightarrow1.47$), and $16.00\%$ ($1.75\rightarrow1.47$), respectively. This highlights the effectiveness of $\xi_{style}$ in style learning. (2) The combination of the $\mathcal{D}_{word}$ and $\mathcal{D}_{line}$ leads to significant improvements in terms of $\text{D}_{\text{CER}}$ and $\text{D}_{\text{WER}}$ without reducing HWD. {This is achieved by employing style-free conditional discriminators, which focus solely on forcing the generator to enhance content readability, without impeding the learning of style.}

\vspace{0.05in}
\noindent \textbf{Qualitative evaluation of style module and content discriminators.}
We conduct visual ablation experiments to further analyze each module in our DiffBrush. As shown in~\cref{fig:ablation}, we observe that the first two basic baselines show clear drawbacks in both style imitation and content readability. Adding our style module significantly improves style reproduction, such as ink color and stroke width, but content readability remains poor. Adding $\mathcal{D}_{word}$ alone improves the character details. However, it struggles to maintain overall content readability, leading to issues like word repetitions and shifts. Employing  $\mathcal{D}_{line}$ solely enhances overall content readability, but character detail issues still remain. In contrast, the best generation results are achieved when both line- and word-level discriminators are used.


\begin{figure}[t]
\begin{center}
\includegraphics[width=0.94\linewidth]{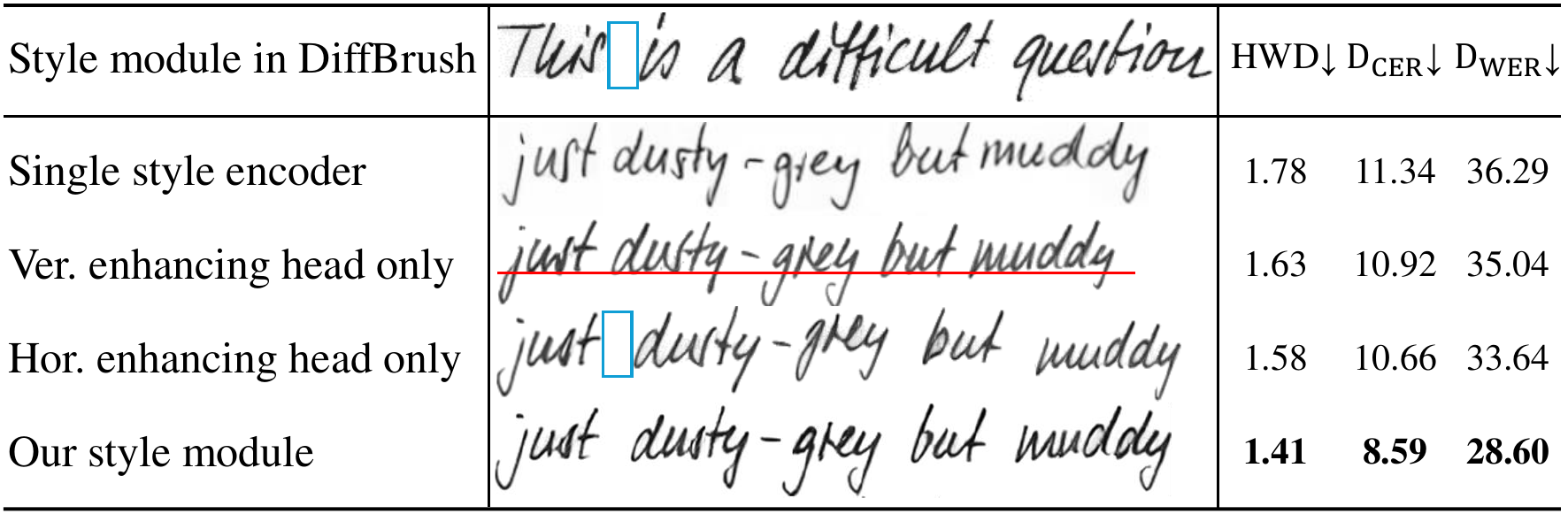}
\end{center}
\vspace{-0.25in}
\caption{{Effect of the horizontal and vertical enhancing heads on IAM test set. Red lines highlight vertical alignment of words, while blue boxes denote the word spacing. ``Single style encoder'' is a basic CNN-Transformer encoder without masking.}}
\label{fig:discussion}
\end{figure}

\begin{figure}[t]
\vspace{-0.05in}
\begin{center}
\includegraphics[width=0.95\linewidth]{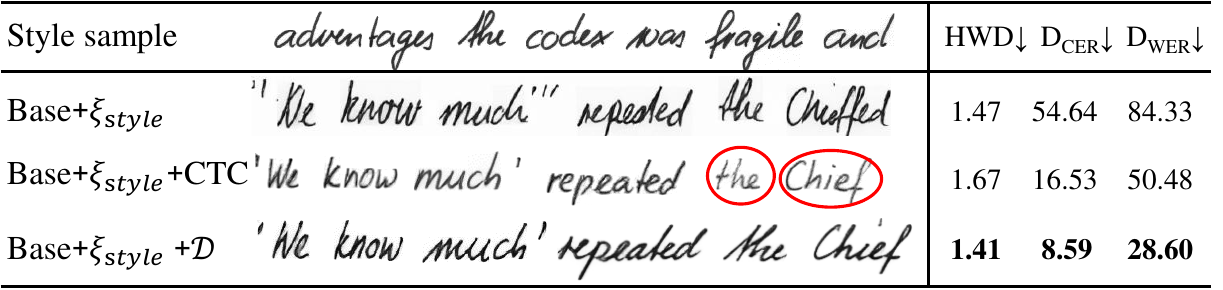}
\end{center}
\vspace{-0.2in}
\caption{{Effect of discriminators $\mathcal{D}$ and CTC recognizer~\cite{graves2006connectionist} under their best trade-off factor. Red circles indicate handwritten texts with simplified style~(\emph{i.e.}, regular texts with standard strokes).}}
\label{fig:discussion_ctc} 
\vspace{-0.25in}
\end{figure}


\vspace{0.05in}
\noindent \textbf{{Discussions about style learning.}}
We conduct ablation study on the style module to analyze the differences between two style-enhancing heads. As shown in~\cref{fig:discussion}, adding either the vertical or horizontal enhancing head improves text-line style quality in terms of HWD.  The vertical head enhances the style imitation ability, particularly in maintaining consistent vertical word alignment. Meanwhile, the horizontal head also improves the style learning, like horizontal spacing patterns. These findings support our motivation that content-masking strategies in different directions help the effective style learning (cf.~\cref{fig:masking}). Finally, it is worth emphasizing that our style features contain complete style information as they are extracted from the entire style reference before any masking (cf.~\cref{fig:DiffBrush}).

\vspace{0.05in}
\noindent \textbf{Discussions on discriminators and CTC.} The quantitative results in~\cref{fig:discussion_ctc} show that incorporating CTC recognizer~\cite{graves2006connectionist} significantly reduces $\text{D}_\text{CER}$ and $\text{D}_\text{WER}$, while also impairing the style evaluation (HWD). Visualization results in~\cref{fig:discussion_ctc} intuitively explain the reasons for this degradation. The CTC version tends to produce texts with simplified styles that differ significantly from style samples. Conversely, our discriminators enhance content readability while preserving style mimicry performance. We provide more experiment details and visual results in Appendix~\ref{Chinese}.

\begin{figure}[t]   

\begin{center}
\includegraphics[width=0.95\linewidth]{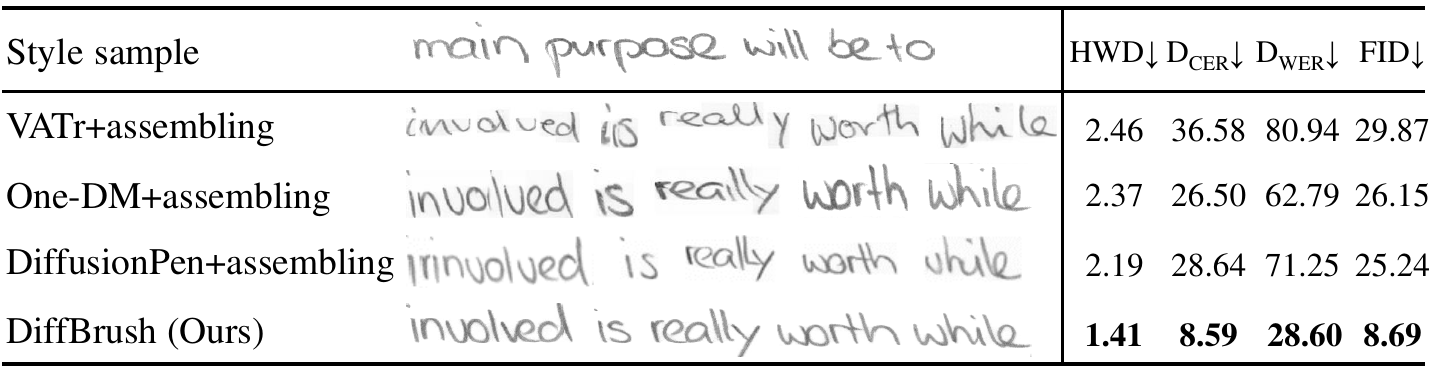}
\end{center}
\vspace{-0.25in}
\caption{Comparisons between directly generated text-lines and text-lines assembled by isolated words on IAM test set.}
\vspace{-0.1in} 
\label{fig:concat}
\end{figure}

\noindent\textbf{{Discussions on directly generated and assembled text-lines.}} We conduct experiments to
demonstrate the superiority of directly generated text lines
over those obtained through concatenation. To this end, we apply the assembling strategy in DiffusionPen~\cite{nikolaidou2024} to enable the official word-level generation baselines to generate text-line image (cf.~\cref{non-text}). The results
in~\cref{fig:concat} demonstrate the superiority of our method over those non-text line methods. More experimental results are provided in~\cref{directly}.

\begin{figure}[t]
\begin{center}
\includegraphics[width=0.95\linewidth]{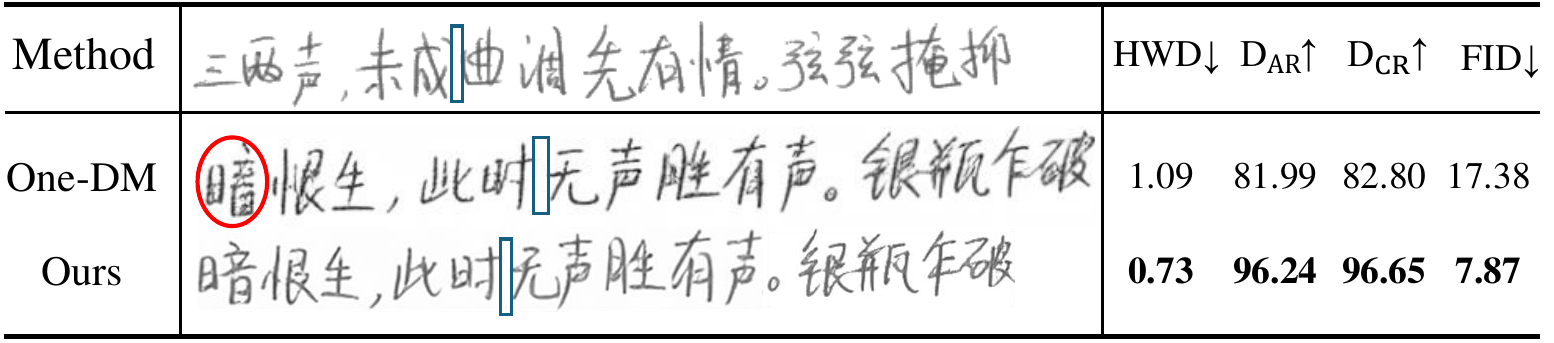}
\end{center}
\vspace{-0.2in}
\caption{{Quantitative and qualitative comparisons with One-DM~\cite{one-dm2024} on Chinese
handwritten text-line generation. The blue boxes highlight the character spacing, while the red circles emphasize the incorrect character structures.}}
\label{fig:application} 
\vspace{-0.25in}
\end{figure}

\vspace{0.05in}
\noindent\textbf{{Applications to Chinese text-line generation.}} We assess DiffBrush's ability to generate Chinese scripts, a challenging task due to thousands of character categories and their complex structures. Both quantitative and qualitative results are provided on~\cref{fig:application}. We observe our DiffBrush effectively handles Chinese handwritten text-lines in terms of style imitation and content fidelity. More experimental details and visualizations are put in~\cref{Chinese}.

\section{Conclusion}
In this paper, we introduce DiffBrush, a novel diffusion model tailored for handwritten text-line generation. To the best of our knowledge, this is among the first exploration of diffusion models for this task. For better style learning and content guidance, we propose a content-decoupled style learning strategy that significantly enhances style imitation and multi-scale content discriminators that supervise textual content at both the line and word levels while preserving style imitation performance. Promising results on three widely-used handwritten datasets verify the effectiveness of our DiffBrush. In the future, we plan to explore its potential to support other generative tasks, such as font generation.

\noindent{\textbf{Acknowledgments}} The research is partially supported by National Natural Science Foundation of China (No.62176093, 61673182), National Key Research and Development Program of China (No.2023YFC3502900), Guangdong Emergency Management Science and Technology Program (No.2025YJKY001).

{
    \small
    \bibliographystyle{ieeenat_fullname}
    \bibliography{main}
}

\clearpage
\appendix
\renewcommand{\thesection}{\Alph{section}}
\maketitlesupplementary

We organize our supplementary material as follows.
\begin{itemize}
\item In~\cref{sec:diffusion}, we review the related work
of diffusion methods in general image generation.
    \item In~\cref{imp}, we describe more implementation details.
    \item In~\cref{content}, we provide more discussions about content evaluation metrics, \emph{i.e,} $\text{D}_\text{CER}$ and $\text{D}_\text{WER}$.
    \item In~\cref{user_study}, we conduct user study experiments.
    \item In~\cref{Chinese}, we put the experimental details and visual results for Chinese handwritten text-line generation.
    \item In~\cref{main_vis}, we provide more ablation experiments: 1) Comparing the content discriminators and CTC recognizer. 2) More visual ablation results of the style module and content discriminators, 3) More visual ablation results of vertical enhancing and horizontal enhancing heads, 4) More ablation results on discriminators architecture, and 5) The effect of masking ratio $\rho$.
    \item In~\cref{non-text}, we provide the implementation details of word-level generation methods.
    \item In~\cref{directly}, we provide more discussions about directly generated text-lines and assembled text-lines.
    \item In~\cref{wier}, we provide more style evaluation results.
     \item In~\cref{Generalization}, we explore the generalization of our DiffBrush to style images with various backgrounds.
    \item In~\cref{downstream}, we explore the downstream application of enriching datasets for training more robust recognizers.
    \item In~\cref{fine-grained}, we discuss the fine-grained style learning.
    \item In~\cref{failure}, we provide failure case analysis.
    \item In~\cref{interpolation}, we present results from visual style interpolation experiments.
    \item In~\cref{English}, we present extensive visual results for English and Chinese handwritten text-line generation.

\end{itemize}

\section{More related work}
\label{sec:diffusion}
\textbf{Image diffusion} Diffusion models such as Denoising Diffusion Probabilistic Model~(DDPM)~\cite{ho2020denoising} and Latent Diffusion Model ~(LDM)~\cite{rombach2022high} have shown great success in image generation. For example, guided diffusion~\cite{dhariwal2021diffusion} and classifier-free diffusion~\cite{ho2022classifier} condition the image synthesis on class labels. Some text-to-image diffusion methods like Stable-diffusion~\cite{rombach2022high} and DALL-E3~\cite{betker2023improving} further employ CLIP~\cite{radford2021learning} to convert text descriptions into comprehensive representations, thereby producing impressive results. Very recently, some methods~\cite{WangZHCZ23,xu2024ufogen} combine adversarial learning with diffusion using a discriminator to enhance generation quality. Unlike these GAN-diffusion approaches that simply distinguish between real and generated images, our two-level content discriminators are specifically designed to provide content supervision at both the line and word levels.

\section{More implementation details}
\label{imp}

\noindent\textbf{Content encoder.}
Following VATr~\cite{pippi2023handwritten} and One-DM~\cite{one-dm2024}, we render the text string $\mathcal{A}$ into Unifont images. The strength of Unifont is its ability to represent all Unicode characters, allowing our method to accept any user-provided string input. We then input the rendered images into a CNN-Transformer content encoder to obtain an informative content feature $Q=\{q_i\}_{i=1}^{L} \in \mathbb{R}^{L\times c}$.

\vspace{0.05in}
\noindent\textbf{Blender.} 
Motivated by One-DM~\cite{one-dm2024}, content $Q$ and style features $S_{ver}$, $S_{hor}$ are fused in the blender with 6 transformer decoder layers~\cite{VaswaniSPUJGKP17,liao2024pptser,liao2025doclayllm,zhang2025mpdrive,peng2024globally,hu2024enhancing,li2024face,lin2024contrastive} across two stages. Initially, $S_{ver}$ serves as the key/value vectors, while $Q$ serves as the query vector that attends to $S_{ver}$ in the first three layers to produce a fused vector. Then, this fused vector becomes a new query vector, attending to $S_{hor}$ in the last three layers as the guiding condition $c \in \mathbb{R}^{L\times c}$.

\vspace{0.05in}
\noindent \textbf{Multi-scale content discriminators.} 
Before being fed into the line content discriminator, the generated image $x_0$ and the content guidance image $I_{line}$ are concatenated along the channel dimension, and the resulting tensor is divided into $n$ segments, as described in~\cref{sec:content}. 
We set $n\small{=}32$ to ensure that each segment approximately covers a single character. Specifically, as mentioned in~\cref{others}, the total width of a text-line image is adjusted to 1024 pixels after data pre-processing. Dividing it into non-overlapping 32 parts yields segments with a width of 32 pixels, closely matching the average character width in the dataset.

Assume the divided segments \( C \in \mathbb{R}^{n \times c \times h \times w} \). A 3D CNN~\cite{tran2015learning} employs sliding window operations across both the spatial dimensions (\emph{i.e.}, \( h \) and \( w \)) and the temporal dimension (\emph{i.e.}, \( n \)) to capture the global contextual information of characters. This representation is then passed to the line content discriminator, which evaluates whether the overall character order in the generated image $x_0$ matches that of the content guidance image $I_{line}$.

\begin{figure}[t]
\begin{center}
\includegraphics[width=0.88\linewidth]{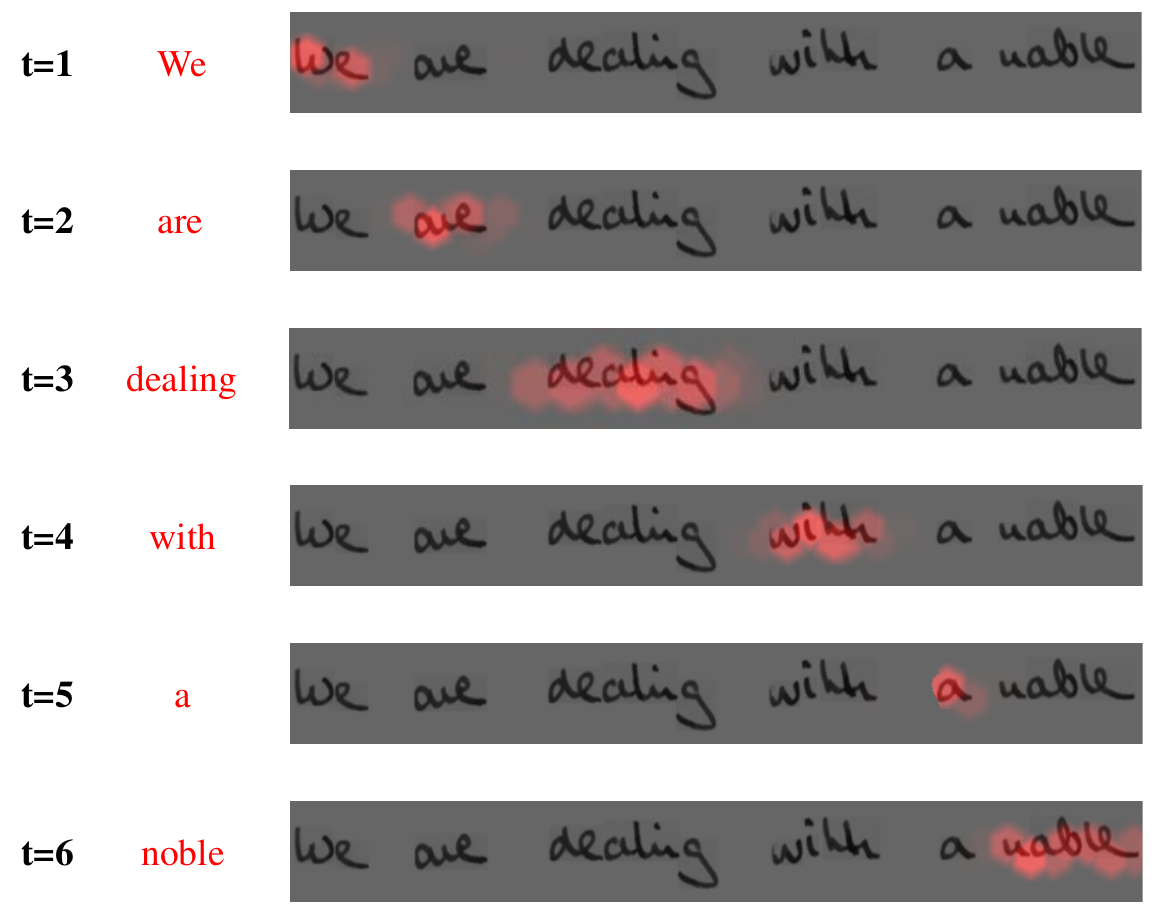}
\end{center}
\vspace{-0.1in}
\caption{Visualization of attention maps for each word in a handwritten text-line image.}
\vspace{-0.2in}
\label{fig:attention_heatmap}
\end{figure}

We pre-train the attention module of the word content discriminator on the training set, enabling it to accurately attend to word positions(cf.~\cref{fig:attention_heatmap}). Its parameters are then frozen during the training of the entire DiffBrush.

\vspace{0.05in}
\noindent{\textbf{Masking strategy.}} Our masking strategy involves randomly masking rows or columns in feature maps (cf.~\cref{fig:sample}). The number of masked elements is determined by the sampling rate $\rho$, which consequently controls both the quantity and size of the masked features. Given $\hat{S} \in \mathbb{R}^{h\times w \times c}$, column-wise masking selects $(w \times \rho)$ tensors of size $h \times 1 \times c$, while row-wise masking selects $(h \times \rho)$ tensors of size $1 \times w \times c$.

\noindent{\textbf{Conditional diffusion generator.}} 
To conserve GPU memory and
accelerate the training time, following Wordstylist~\cite{2023wordstylist} and One-DM~\cite{one-dm2024}, we streamline the U-Net by reducing the number
of ResNet~\cite{xie2025random} blocks and attention heads and take the diffusion process into the latent space. Specifically, we adopt a powerful, pre-trained Variational
Autoencoder (VAE) of Stable Diffusion (1.5) to convert the image into latent space. During the training phase, we freeze the parameters of VAE and set $T = 1000$ steps, and forward process variances are set to constants increasing linearly from $\beta_{1} = 10^{-4}$
to $\beta_{T} = 0.02$.

\vspace{0.05in}
\noindent{\textbf{More training details.}} 
The proposed conditional diffusion generator $\mathcal{G}$ and the multi-scale discriminators $\mathcal{D}$ engage in an adversarial learning process: $\mathcal{G}$ seeks to synthesize realistic images that $\mathcal{D}$ cannot distinguish from real ones based on content, while $\mathcal{D}$ assess the content at both the line and word scales. The readability of the generated images improves through two adversarial losses, $\mathcal{L}_{line}$ and $\mathcal{L}_{word}$, which further enhances generation quality in terms of content accuracy. In summary, the overall training objectives for the conditional diffusion generator, 
and the multi-scale discriminators are defined as:
\begin{equation}
\begin{aligned}
\mathcal{L}_{\mathcal{G}}=\mathcal{L}_{diff} + \mathcal{L}_{style} +  \lambda\mathcal{L}_{content},
\end{aligned}
\end{equation}
\vspace{-0.2in}
\begin{equation}
\begin{aligned}
\mathcal{L}_{\mathcal{D}}= -\mathcal{L}_{content},
\end{aligned}
\end{equation}

Our DiffBrush is trained with 800 epochs, as described in~\cref{others}. During the first 750 epochs, we optimize the model using only $\mathcal{L}_{ver}$, $\mathcal{L}_{hor}$ and $\mathcal{L}_{diff}$. In the final 50 epochs, we retain these loss functions and further introduce $\mathcal{D}_{line}$ and $\mathcal{D}_{word}$ to enhance the readability of the generated images $x_0$. Specifically, we use the conditional diffusion generator to perform 5 denoising steps, generating a handwritten image $x_0$ with coarse content structure. We then input $x_0$ to the multi-scale content discriminators to obtain content supervision at both line and word levels. The total training time for our DiffuBrush is approximately 4 days.

\section{More discussions about $\text{D}_\text{CER}$ and $\text{D}_\text{WER}$}
\label{content}
To better evaluate the content quality of generated results, we use $\text{D}_\text{CER}$ and $\text{D}_\text{WER}$ as content evaluation metrics, following the recent works~\cite{rethinking,nikolaidou2024,one-dm2024}. It is worth emphasizing that $\text{D}_\text{CER}$ is also referred to as $\text{HTG}_{\text{HTR}}$ in ~\cite{rethinking}. The difference lies in the fact that $\text{HTG}_{\text{HTR}}$ measures character error rate (CER) at the word level, whereas $\text{D}_\text{CER}$ measures it at the text-line level.

More specifically, the implementation details of $\text{D}_\text{CER}$ and $\text{D}_\text{WER}$ are as follows: 1) Each method is employed to generate a complete training set, which is then used to train an OCR system~\cite{retsinas2022best}. The system is built on a CNN-LSTM architecture with a CTC loss~\cite{graves2006connectionist,huang2021context,zhuang2021new}. 2) The character error rate (CER) and word error rate (WER) are evaluated on the real test set, aiming to achieve recognition performance as close as possible to that obtained with a real training set.


As noted in~\cite{rethinking,nikolaidou2024}, the intuition behind this experiment is that a handwriting generation method achieving or exceeding the performance of the real IAM dataset demonstrates two crucial abilities: 1) The generated handwritten text images have accurate content. 2) The generated samples exhibit diverse styles. 
Although the first criterion is crucial, focusing only on it while overlooking the second criterion can lead to biased evaluations. For instance, if a generation method favors producing easily recognizable handwritten texts with simplified styles (cf. red circles in~\cref{fig:ablation_D_ctc}), it might achieve high content accuracy. However, such low-diversity results are not satisfactory to us. To address this, we use $\text{D}_\text{CER}$ and $\text{D}_\text{WER}$ to provide a more comprehensive evaluation of the content quality in the generated samples, since  $\text{D}_\text{CER}$ and $\text{D}_\text{WER}$ simultaneously take the aforementioned two factors into account.



\section{User studies}
\label{user_study}
\begin{figure}[t]
\begin{center}
\includegraphics[width=0.8\linewidth]{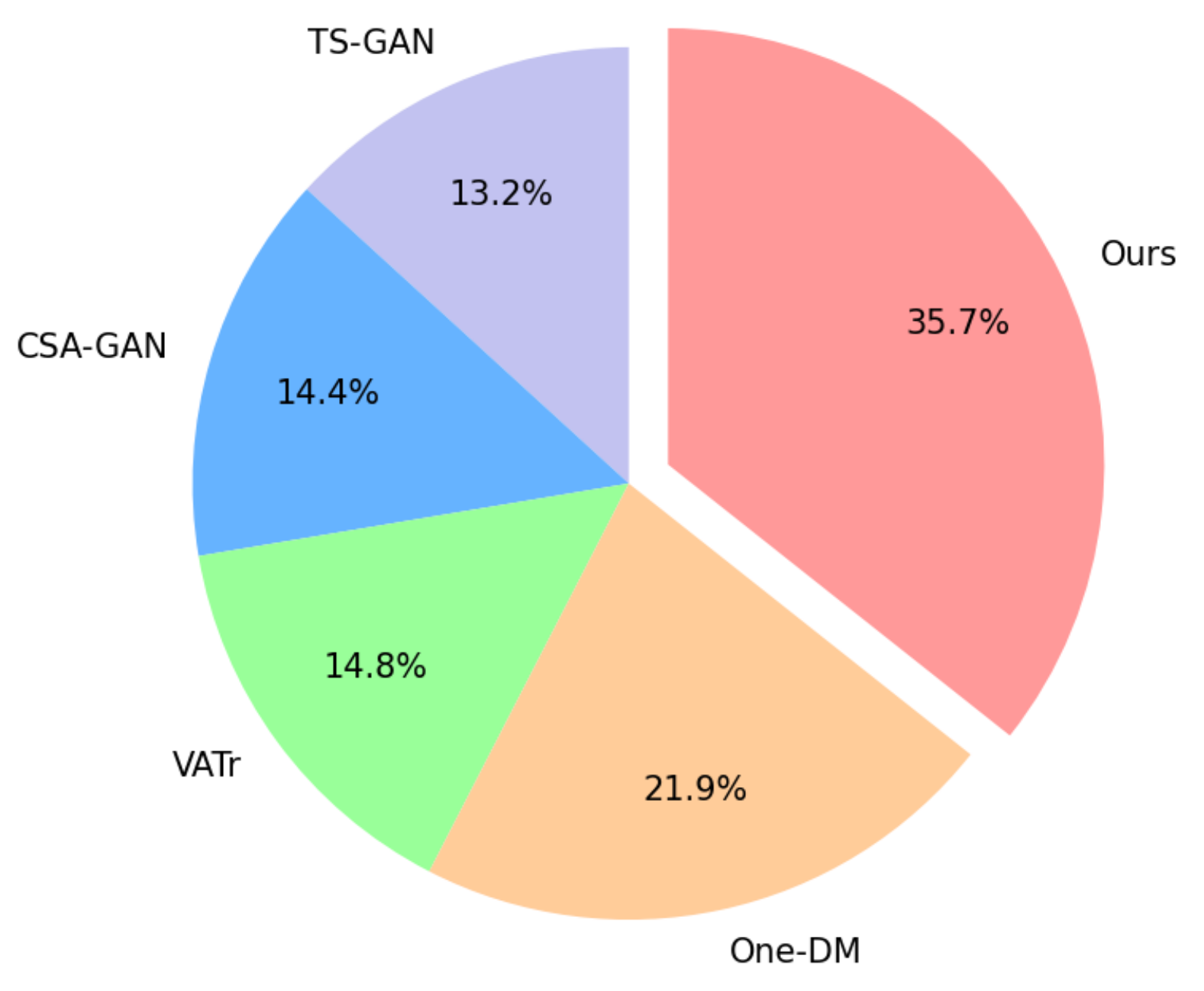}
\end{center}
\vspace{-0.1in}
\caption{User preference study with a comparison to state-of-the-art methods on handwritten text-line generation.}
\label{fig: user preference}
\end{figure}

\textbf{User preference study.} 
We invite human participants with postgraduate education backgrounds to evaluate the visual quality of synthesized handwritten text images, focusing on style imitation. The generated samples are from our method and other state-of-the-art approaches. In each round, we randomly select a writer from the IAM dataset and use their handwritten text-line sample as style guidance, along with identical text as content guidance, to direct all methods in generating candidate samples. Participants are presented with one text-line from the exemplar writer as a style reference and multiple candidates generated by different methods. They are asked to select the candidate that best matches the reference in style. This process is repeated 30 times, yielding 900 valid responses from 30 volunteers. As shown in~\cref{fig: user preference}, our method receives the most user preferences, demonstrating its superior quality in style imitation.

\noindent{\textbf{User plausibility study.}} 
We conduct a user plausibility study to assess whether the text-line images generated by DiffBrush are indistinguishable from real handwriting samples. In this study, participants are first shown 30 examples of authentic handwritten text-line samples. They are then asked to classify each image they see as either real or synthetic, with the images being randomly selected from both genuine samples and those generated by our method. In total, 30 participants provide 900 valid responses. The results, shown as a confusion matrix in~\cref{plausi}, report a classification accuracy close to $50\%$, suggesting the task becomes equivalent to random guessing. This indicates that text-line images generated by our method are nearly indistinguishable from real samples.

\begin{table}[t]
    \centering 
    \scalebox{1.00}{
    \begin{tabular}{l cc c}
     \toprule  
      \multirow{2}*{Actual}   & 
         \multicolumn{2}{c}{Predicted}
         &  \multirow{2}*{\makecell{Classification \\ Accuracy }} \\
         \cmidrule{2-3}
     
      ~ &  Real & Fake    & ~   \\
        \midrule
        
        Real  &  27.22  & 22.78   &  \multirow{2}*{49.11}  \\
        Fake & 28.11 & 21.89 & ~ \\
        \bottomrule
    \end{tabular}}
    \caption{Confusion matrix$\left(\%\right)$ from the user plausibility study. The classification accuracy of $49.11\%$ suggests that users struggle to differentiate between handwritten text-line images generated by our DiffBrush and real ones.}
    \label{plausi}

\end{table}

\section{Chinese handwritten text-line generation}
\label{Chinese}
In this section, we evaluate DiffBrush's capability to generate scripts with thousands of character categories and complex character structures, such as Chinese. For this purpose, we perform experiments on the widely used Chinese handwritten text-line dataset CASIA-HWDB (2.0–2.2)~\cite{liu2011casia}. 

\noindent{\textbf{Dataset.}} CASIA-HWDB (2.0-2.2) consists of 52,230 Chinese text-lines belonging to 1,019 different writers. Following the standard split of CASIA-HWDB (2.0–2.2), we use text-lines from 816 writers for training and remaining 203 writers for testing. As mentioned in~\cref{imp}, since Unifont~\cite{pippi2023handwritten,one-dm2024} can encode all Unicode characters, we still employ it to convert Chinese strings into textual content images. Similarly, in our experiments, all images are adjusted to 64 $\times$ 1024 pixels.

\noindent{\textbf{Evaluation Metrics.}} Similarly, we use HWD and FID~\cite{heusel2017gans,gong2025monocular} to evaluate style imitation and visual quality of generated handwritten images, respectively. We still use the OCR system~\cite{retsinas2022best} to measure the content quality of generated results. Unlike English handwriting generation, we evaluate the OCR system's recognition performance on a real test set using two widely adopted metrics in Chinese text-line recognition~\cite{su2009off,wang2011handwritten,chen2023improved}: Accuracy Rate (AR) and Correctness Rate (CR). Similarly, we name them $\text{D}_\text{AR}$ and $\text{D}_\text{CR}$.


\begin{table}[t]
\centering
\scalebox{0.9}{\begin{tabular}{lcccc}
\toprule
                DiffBrush & $\lambda$ & HWD $\downarrow$ & $\text{D}_{\text{CER}} \downarrow$ & $\text{D}_{\text{WER}} \downarrow$ \\ 
                \midrule
                \multirow{4}{*}{w/ CTC~\cite{graves2006connectionist}} & 1 & 2.01 & 19.62 & 52.64 \\ 
                & 0.1 & \textbf{1.67} & \textbf{16.53} & \textbf{50.48}  \\ 
                & 0.01 & 1.69 & 17.12 & 51.34 \\ 
                & 0.001 & 1.68 & 17.29 & 50.72 \\ 
\toprule

\multirow{3}{*}{w/ Discriminators} & 0.1 & 1.43 & 9.87 & 31.96  \\ 
                & 0.05 & \textbf{1.41} & \textbf{8.59} & \textbf{28.60} \\ 
                & 0.01 & 1.45 & 10.40 & 32.54  \\
                \bottomrule
\end{tabular}}
\caption{Quantitative ablation results for the CTC recognizer and content discriminator variants of our DiffBrush on IAM test set. $\lambda$ denotes the trade-off factor (cf.~\cref{eq1} in~\cref{sec:overall}).}
\label{content_ablation}
\vspace{-0.2in}
\end{table}

\noindent{\textbf{Qualitative comparison.}} We provide qualitative results in~\cref{fig:Chinese_1} and~\cref{fig:Chinese_2}. To ensure fair comparisons, both DiffBrush and One-DM~\cite{one-dm2024} are conditioned on the same text contents and style samples. We observe that the handwritten text lines generated by our DiffBrush (rows of ``Ours") exhibit styles most similar to the reference samples, particularly in terms of character spacing and ink color, while preserving accurate character structures. 

\section{More ablation results}
\label{main_vis}
\subsection{More ablation on discriminators and CTC}
\begin{figure}[t]
\begin{center}
\includegraphics[width=0.95\linewidth]{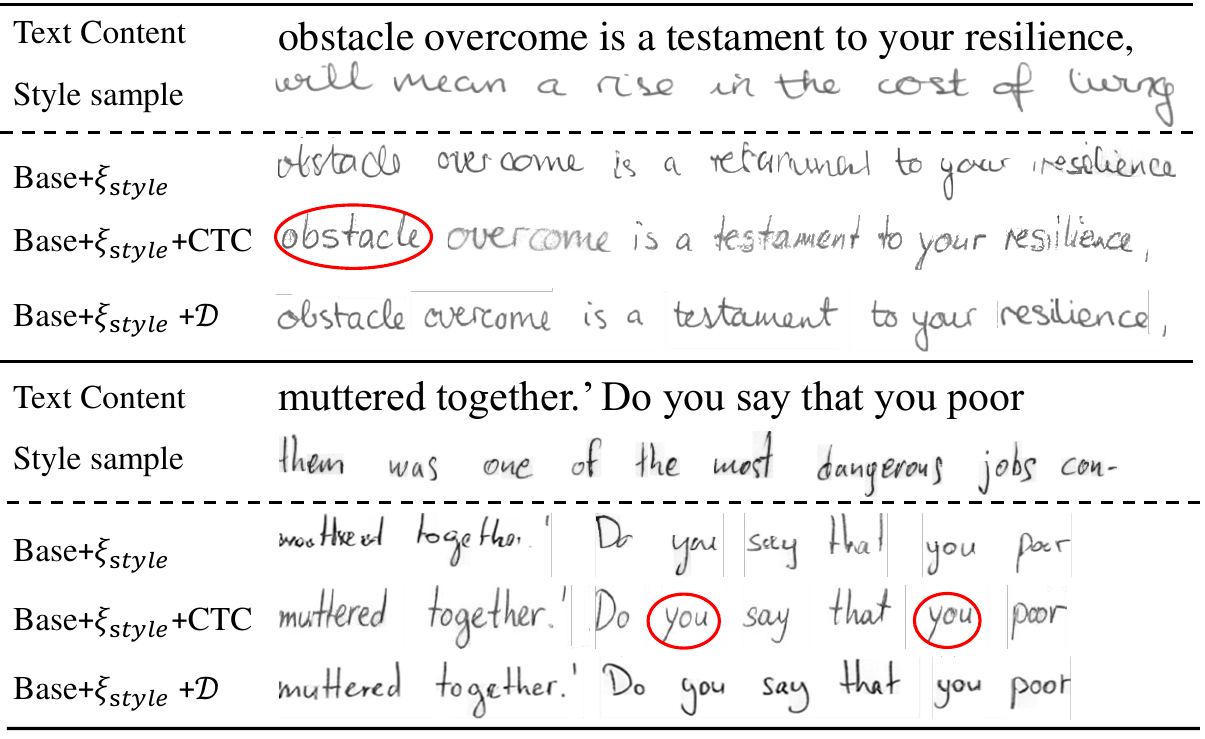}
\end{center}
\vspace{-0.1in}
\caption{Qualitative comparisons between the discriminators $\mathcal{D}$ and the CTC recognizer~\cite{graves2006connectionist}. Red circles indicate handwritten texts with simplified style~(\emph{i.e.}, regular texts with standard strokes).}
\label{fig:ablation_D_ctc}
\end{figure}

To further evaluate the impact of the proposed content discriminators, we replace the multi-scale content discriminator in DiffBrush with the standard CTC recognizer~\cite{graves2006connectionist} adopted in One-DM~\cite{one-dm2024} and conduct experiments to compare them. The quantitative results in~\cref{content_ablation} show that our discriminator version performs better. In~\cref{fig:ablation_D_ctc}, we further visualize their best $\lambda$ results on IAM test set. We observe that CTC tends to simplify handwriting styles, while the discriminator version better preserves reference styles.

\subsection{More ablation results of $\xi_{style}$ and \texorpdfstring{$\mathcal{D}$}{D}}
\begin{figure*}[t]
\begin{center}
\includegraphics[width=0.85\linewidth]{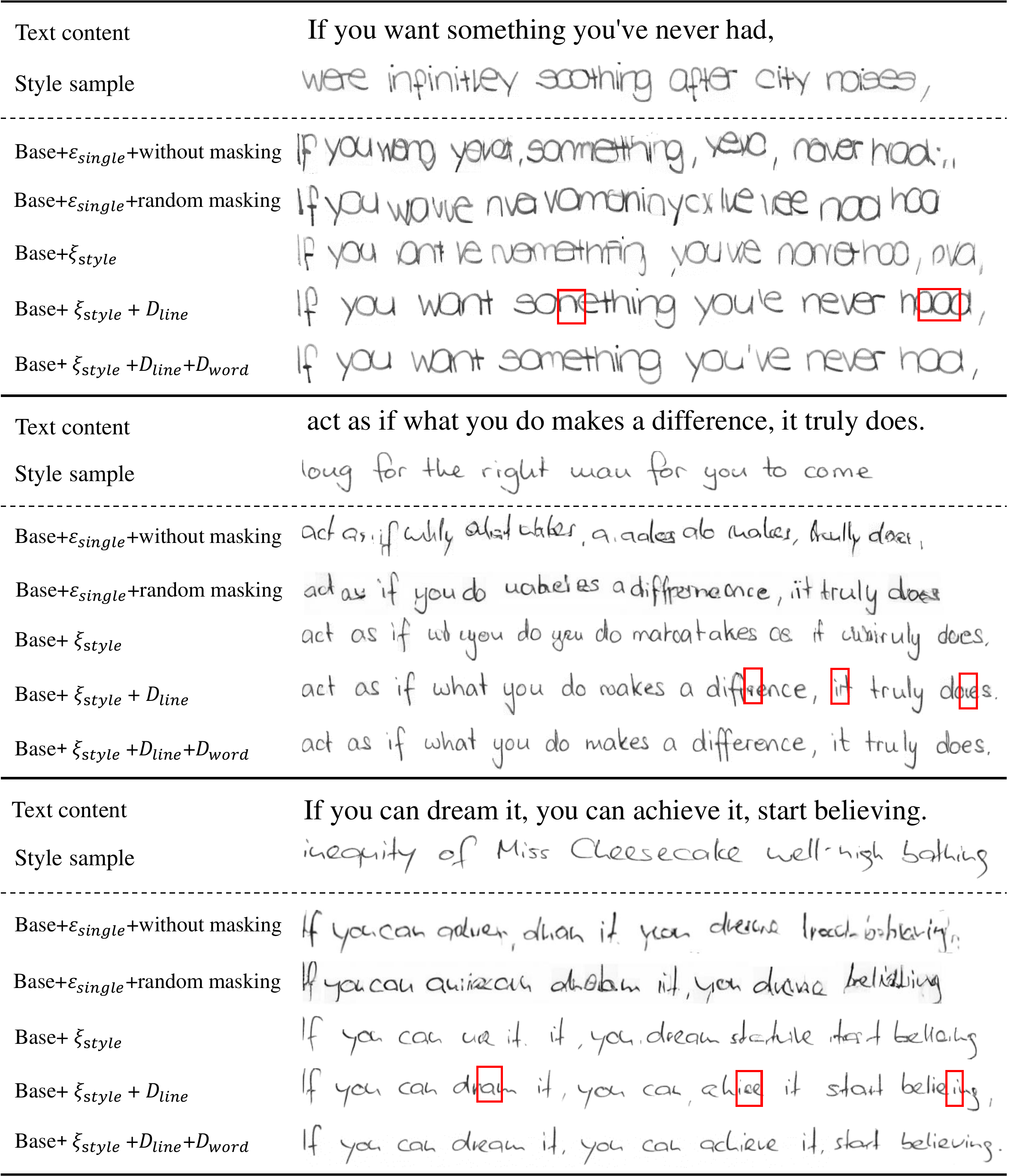}
\end{center}
\vspace{-0.1in}
\caption{More visual ablation results of the style module $\xi_{style}$ and content discriminators (\emph{i.e.}, $\mathcal{D}_{line}$ and $\mathcal{D}_{word}$). $\varepsilon_{single}$ denotes a single CNN-Transformer style encoder. The red boxes highlight failures of structure preservation.}
\label{fig:ablation_main}
\end{figure*}

In~\cref{fig:ablation_main}, we provide more qualitative results of the style module $\xi_{style}$ and content discriminators (\emph{i.e.}, $\mathcal{D}_{line}$ and $\mathcal{D}_{word}$) on IAM dataset. From these results, we can observe that $\varepsilon_{single}$ versions exhibit notable flaws in both content accuracy and style imitation, including inconsistencies in stroke color, thickness, and word spacing compared to the style samples.  Incorporating $\xi_{style}$ significantly improves style-related flaws; however, the readability of the generated text remains unsatisfactory. The introduction of $\mathcal{D}_{line}$ ensures that the overall character order in the generated text closely matches the content reference, greatly improving content accuracy. Nevertheless, certain character details remain imperfect. After adding $\mathcal{D}_{word}$, incorrect character structures are effectively corrected, thus further enhancing the readability of the generated text-line.

\subsection{Architecture ablation on discriminators}
\label{discriminators}
Our line content discriminator assesses global character order, while the word discriminator verifies local text accuracy. To this end: 1) A 3D CNN processes segmented character fragments to learn global character context.  2) A 2D CNN focuses on the content information of the attended words. In~\cref{tab:3d cnn}, an ablation study on IAM dataset show that: 1) Replacing $\mathcal{D}_{line}$ with a 2D CNN significantly increases $\text{D}_{\text{CER}}$ and $\text{D}_{\text{WER}}$,  highlighting the limitations of using 2D CNNs for full text lines. 2) Switching $\mathcal{D}_{word}$ to a 3D CNN yields no significant change, as a 2D CNN is sufficient for attended words with fewer characters.

\begin{table}[t]
\centering
\scalebox{0.9}{\begin{tabular}{cccc}
                \toprule
                Discriminators & HWD $\downarrow$ & $\text{D}_{\text{CER}} \downarrow$ & $\text{D}_{\text{WER}} \downarrow$ \\ 
                \midrule
                2D $\mathcal{D}_{line}$ + 2D $\mathcal{D}_{word}$ & 1.44 & 13.18 & 39.64 \\ 
                3D $\mathcal{D}_{line}$ + 3D $\mathcal{D}_{word}$ & 1.43 & 9.58 & 30.27 \\ 
                Ours (3D $\mathcal{D}_{line}$ + 2D $\mathcal{D}_{word}$) & \textbf{1.41} & \textbf{8.59} & \textbf{28.60} \\ 
                \bottomrule
            \end{tabular}}
\vspace{-0.05in}
\caption{{Architecture ablation results on content discriminators.}}
\label{tab:3d cnn}
\end{table}

\subsection{More ablation results on style learning}
\begin{figure*}[t]
\begin{center}
\includegraphics[width=0.85\linewidth]{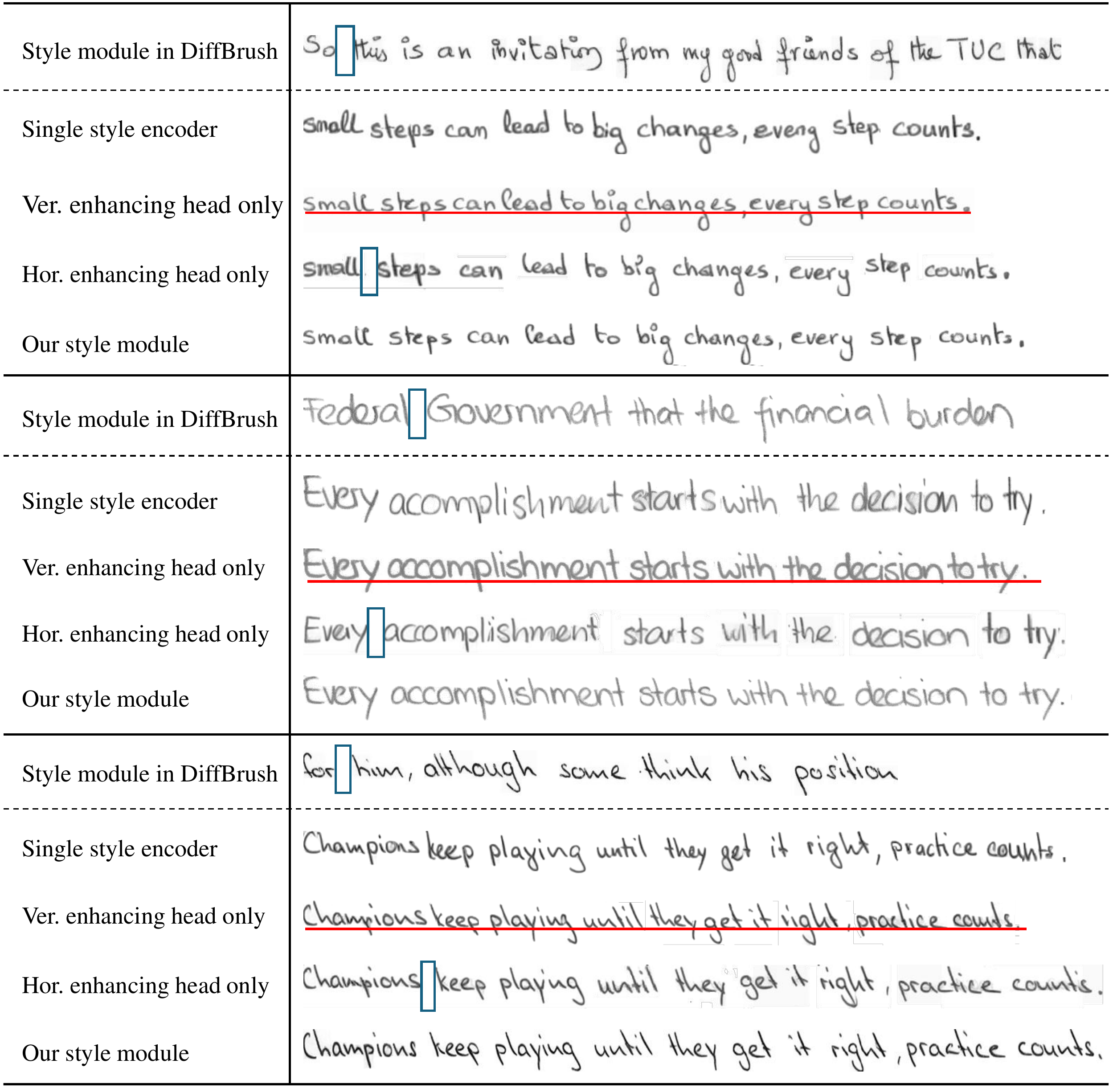}
\end{center}
\vspace{-0.1in}
\caption{More visual ablation results of vertical enhancing and horizontal enhancing heads. The red lines indicate alignment of words along the vertical axis, while blue boxes indicate word spacing.}
\label{fig:more_discussion}
\end{figure*}

More visual ablation results are provided in~\cref{fig:more_discussion} to analyze the effect of vertical enhancing and horizontal enhancing heads. The findings indicate that incorporating the vertical enhancing head enhances the model's style imitation capabilities, notably in maintaining vertical alignment between words. Similarly, adding the horizontal enhancing head also improves the learning of style patterns, especially in preserving horizontal word spacing. The integration of both heads is crucial for our DiffBrush to produce high-quality results that faithfully replicate the writing styles of the reference samples.

\begin{table}[t]
\centering
\scalebox{1.00}{
\begin{tabular}{cccc}
                \toprule
                Masking ratio & HWD $\downarrow$ & $\text{D}_{\text{CER}} \downarrow$ & $\text{D}_{\text{WER}} \downarrow$ \\ 
                \midrule
                0 & 1.72 & 56.52 & 87.15 \\ 
                0.25 & 1.64 & 56.48 & 86.99 \\ 
                0.50 & \textbf{1.47} & \textbf{54.64} & \textbf{84.33} \\ 
                0.75 & 1.58 & 55.29 & 88.34 \\ 
                \bottomrule
            \end{tabular}}
\caption{Effect of the masking ratio $\rho$. The results are derived from our DiffBrush without using multi-scale content discriminators.}
\label{tab:masking}
\end{table}

\subsection{Analysing the effect of masking ratio}
\label{masking}
We analyze $\rho$ by removing content discriminators from DiffBrush. As shown in~\cref{tab:masking}, $\rho=0.5$ yields the best performance on IAM test set.

\section{Comparisons with non-text line methods} 
\label{non-text}

As shown in~\cref{fig:line_samples}, directly \textbf{assembling isolated words} from official word-level generation models leads to unnatural and low-quality text-line results. We thus retrain these non-text line methods (\emph{i.e.,} VATr~\cite{pippi2023handwritten}, DiffusionPen~\cite{nikolaidou2024} and One-DM~\cite{one-dm2024}) on text-line dataset to enable them to directly generate text lines for fair comparisons. It is worth emphasizing that the official text-line generation scheme in DiffusionPen~\cite{nikolaidou2024} also employs an \textbf{assembly-based strategy}. This involves resizing each generated word image to ensure consistent character width, followed by a concatenation operation with a fixed space.

\section{Discussion on generated and assembled line}
\label{directly}
We further conduct more experiments on IAM dataset to demonstrate the superiority of directly generated text lines. We utilize our DiffBrush to directly generate handwritten text-line images. For the assembled lines, we employ One-DM~\cite{one-dm2024} to produce isolated words, which are then concatenated into text lines using statistical methods. More specifically, given a text-line style reference, we first employ the Otsu algorithm~\cite{otsu1975threshold} to compute a binary
mask of the text-line image, effectively separating words from the background. We then calculate the average spacing $m$ between words and determine whether the centroids of the words are aligned along a horizontal or skewed line. With this statistical information, we concatenate the synthesized words from One-DM, ensuring that the spacing between words is $m$,  and that the centroids of the words maintain the consistent vertical alignment patterns as the text-line reference.

\begin{table}[t]
\centering
\scalebox{0.8}{\begin{tabular}{lcccc}
                \toprule
                Method & HWD $\downarrow$ & $\text{D}_{\text{CER}} \downarrow$ & $\text{D}_{\text{WER}} \downarrow$ & FID$\downarrow$\\ 
                \midrule
               One-DM + post-processing & 2.17 & 24.81 & 62.08 & 23.92 \\ 
                DiffBursh (Ours) & \textbf{1.41} & \textbf{8.59} & \textbf{28.60} & \textbf{8.69} \\ 
                \bottomrule
            \end{tabular}}
\caption{Quantitative comparisons between directly generated and assembled text-lines on the IAM test set.}
\label{tab:quanti-ass}
\end{table}

\begin{figure}[t]
\begin{center}
\includegraphics[width=0.9\linewidth]{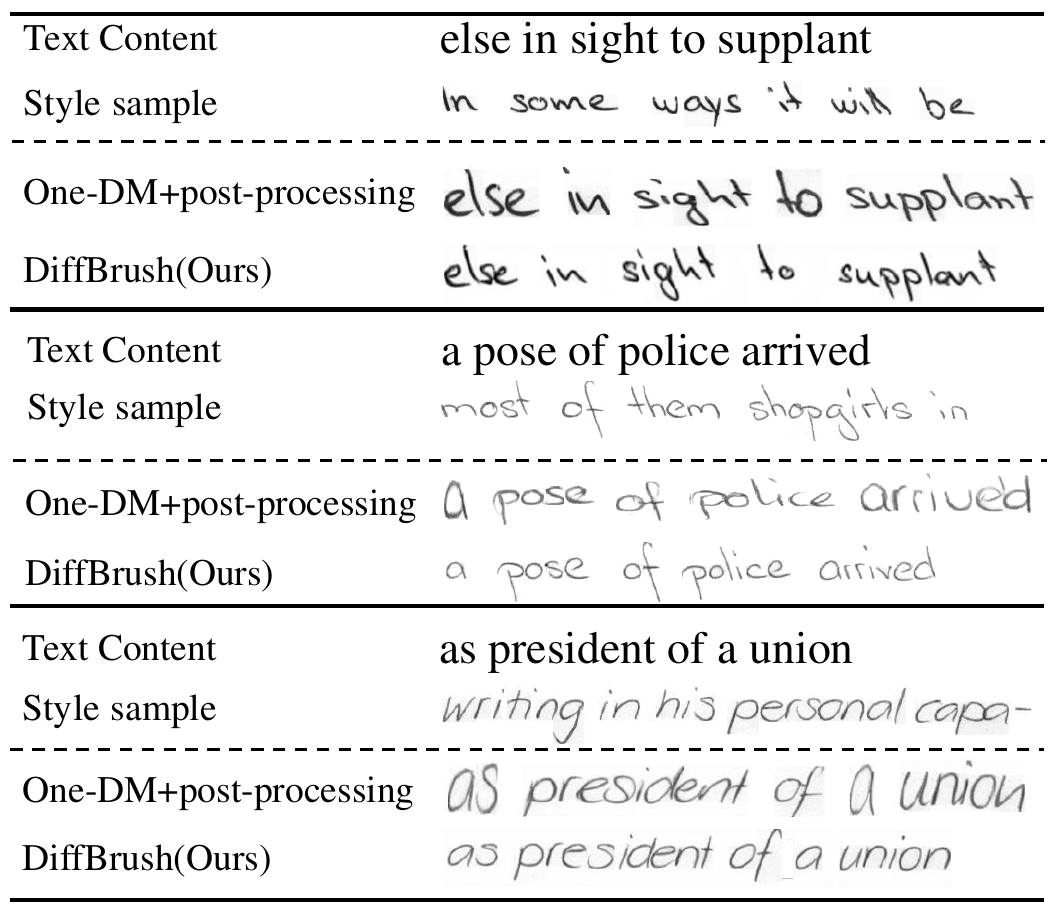}
\end{center}
\vspace{-0.2in}
\caption{Qualitative comparisons between directly generated and assembled text-lines on the IAM test set.}
\label{fig:ass}
\end{figure}

Quantitative results in~\cref{tab:quanti-ass} indicate that the text-lines generated by our DiffBrush significantly outperform assembled text-lines in terms of style evaluation (HWD), content evaluation ($\text{D}_{\text{CER}}$ and $\text{D}_{\text{WER}}$), visual quality evaluation (FID). These results demonstrate the advantages of direct generation. From qualitative results in~\cref{fig:ass}, we can observe that our directly generated text-lines exhibit more consistent stroke colors, uniform character sizes, and vertical word alignment patterns that more closely match the style samples. These findings further underscore the superiority of directly generating text lines.

\section{Discussions on text-line style evaluation} 
\label{wier}
We did not use the writer classifier from previous works~\cite{gan2022higan+,zdenek2023handwritten} because the models were designed for evaluating \textbf{word-level text style} and the IAM split details for training the classifier were missing, making fair and reproducible evaluation difficult. Instead, we adopt the open-source HWD metric~\cite{PippiQCC23}, which offers two advantages: (1) it ensures reproducibility, and (2) it is pre-trained on large-scale handwritten data and proven effective in \textbf{text-line style} evaluation. Following the comment, we further include WIER~\cite{gan2022higan+,zdenek2023handwritten} for style evaluation. To this end, we randomly split the standard IAM test set into 80\% for classifier~\cite{gan2022higan+} training and 20\% for validation. The best classifier is then used to evaluate the generation results. As shown in \cref{style_acc}, our DiffBrush continues to achieve the best style imitation performance.

\section{Generalization to more style backgrounds}
\label{Generalization}
\begin{figure*}[t]
\begin{center}
\includegraphics[width=0.9\linewidth]{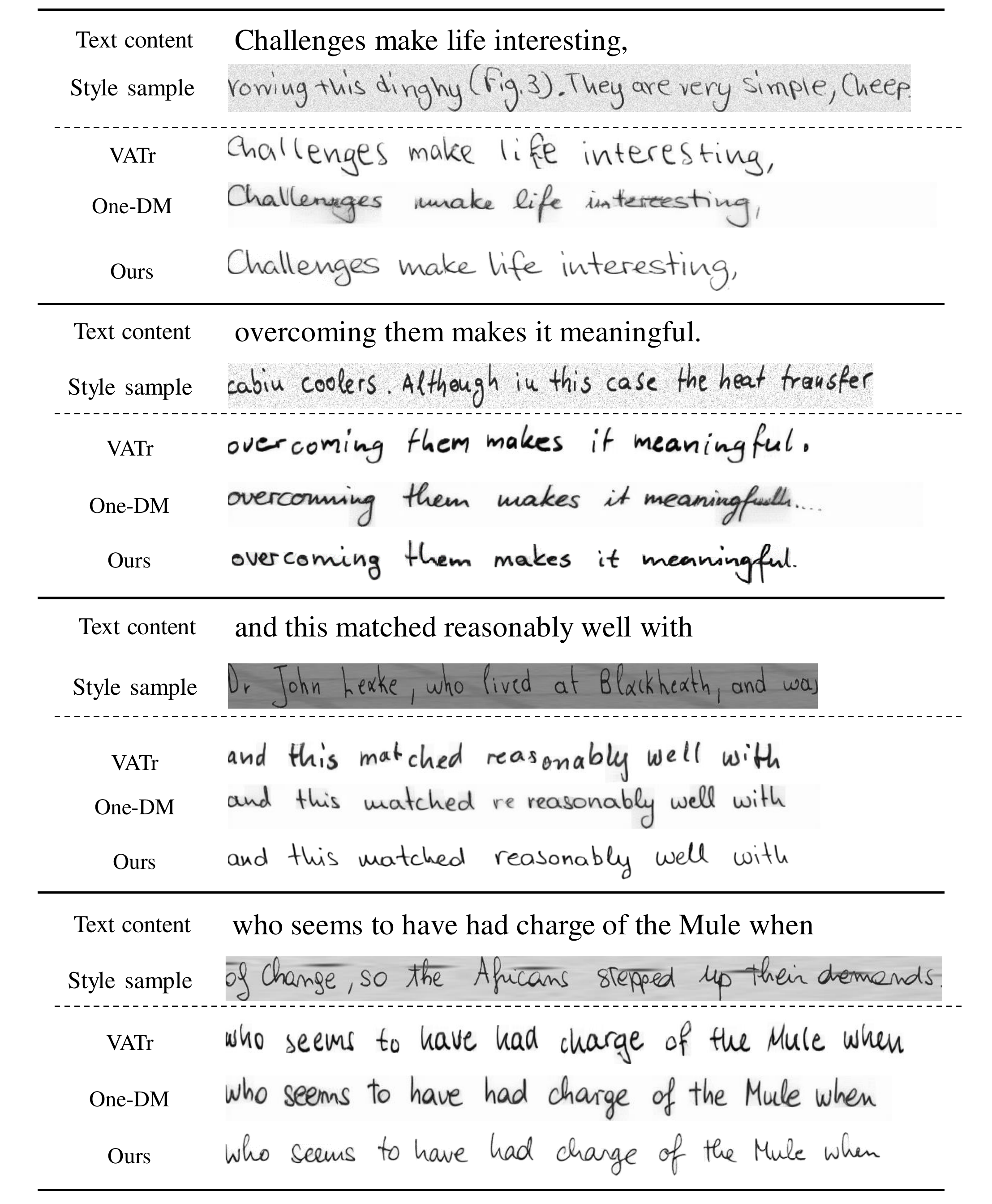}
\end{center}
\vspace{-0.1in}
\caption{Generated handwritten text-line images conditioned on style samples with more complex and realistic backgrounds.}
\label{fig:addbg_1}
\end{figure*}

\begin{figure*}[t]
\begin{center}
\includegraphics[width=0.9\linewidth]{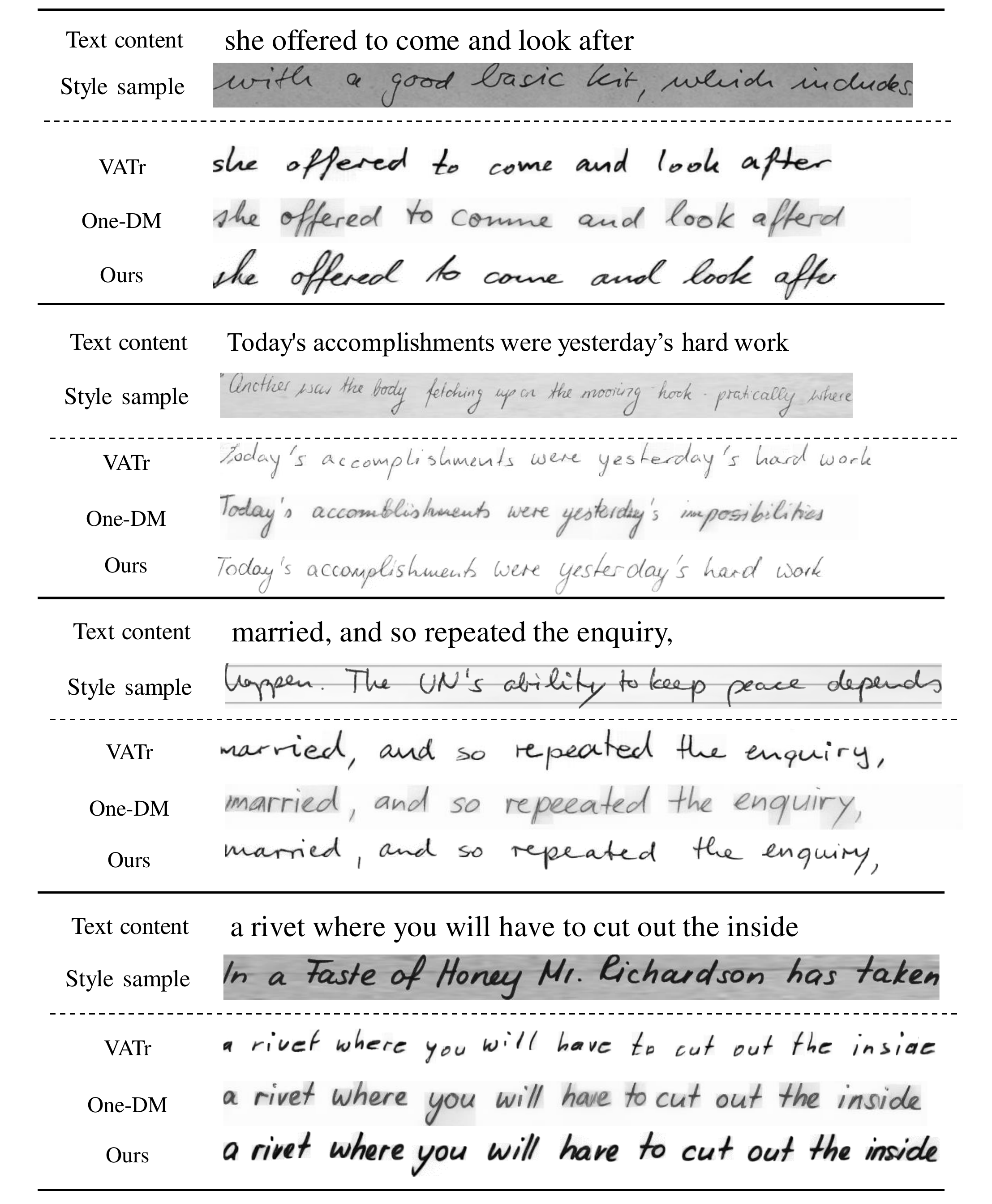}
\end{center}
\vspace{-0.1in}
\caption{Generated handwritten text-line images conditioned on style samples with more complex and realistic backgrounds.}
\label{fig:addbg_2}
\end{figure*}

To assess whether DiffBrush can effectively generalize to different style backgrounds, we condition it on eight complex and realistic backgrounds. The generated results are shown in~\cref{fig:addbg_1} and~\cref{fig:addbg_2}. We find that our DiffBrush still generates high-quality handwritten text-line images, further demonstrating the robustness of our DiffBrush.

\section{Application for training robust recognizer}
\label{downstream}
A key application of handwritten text-line generation models is to enrich the training dataset, facilitating the training of more robust recognizers. To this end, we combine the IAM training set generated by various methods with the real training set to create a new mixed dataset. We then train an OCR system~\cite{retsinas2022best} using this mixed dataset and report its performance on the real IAM test set. We present the quantitative results in~\cref{enrich}. These results clearly show that the additional synthetic data contributes to improving the recognizer's performance. Among all methods, our approach achieves the greatest performance improvement, with an improvement rate of $20.07\%$.

\begin{table}[t]
    \centering
        \scalebox{0.6}{\begin{tabular}{lcccccc}
            \toprule
            Method &  TS-GAN & CSA-GAN  & VATr &  DiffusionPen &  One-DM & Ours \\ 
            \midrule
            WIER ($\%$) $\downarrow$ & 96.03 & 82.14 & 76.96 & 73.85 & 70.92 & \textbf{59.77} \\ 
            \bottomrule
        \end{tabular}}

        \vspace{-0.1in}
        \caption{Style evaluation on the IAM test set.}
        \label{style_acc}
        \vspace{-0.1in}
\end{table}

\begin{table}[t]
\centering
\scalebox{0.85}{
\begin{tabular}{lccc}
\toprule
Training Data        & CER $\downarrow$ & WER $\downarrow$ & Improve. (\%) $\uparrow$\\
\midrule
Real                 & 5.78             & 21.76            & -                     \\
CSA-GAN + Real       & 5.39             & 19.89            & 6.74                  \\
VATr + Real          & 5.08             & 19.31            & 12.11                 \\
One-DM + Real        & 4.99             & 18.51            & 13.67                 \\
DiffBrush (Ours) + Real & \textbf{4.62}  & \textbf{16.86}   & \textbf{20.07}        \\
\bottomrule
\end{tabular}}
\caption{Handwritten text-line recognition on different training data. Improvement rate refers to CER performance gain achieved by incorporating synthetic data into the training process compared to using only the real training set.}
\label{enrich}
\end{table}

\begin{figure}[t]
\begin{center}
\includegraphics[width=0.95\linewidth]{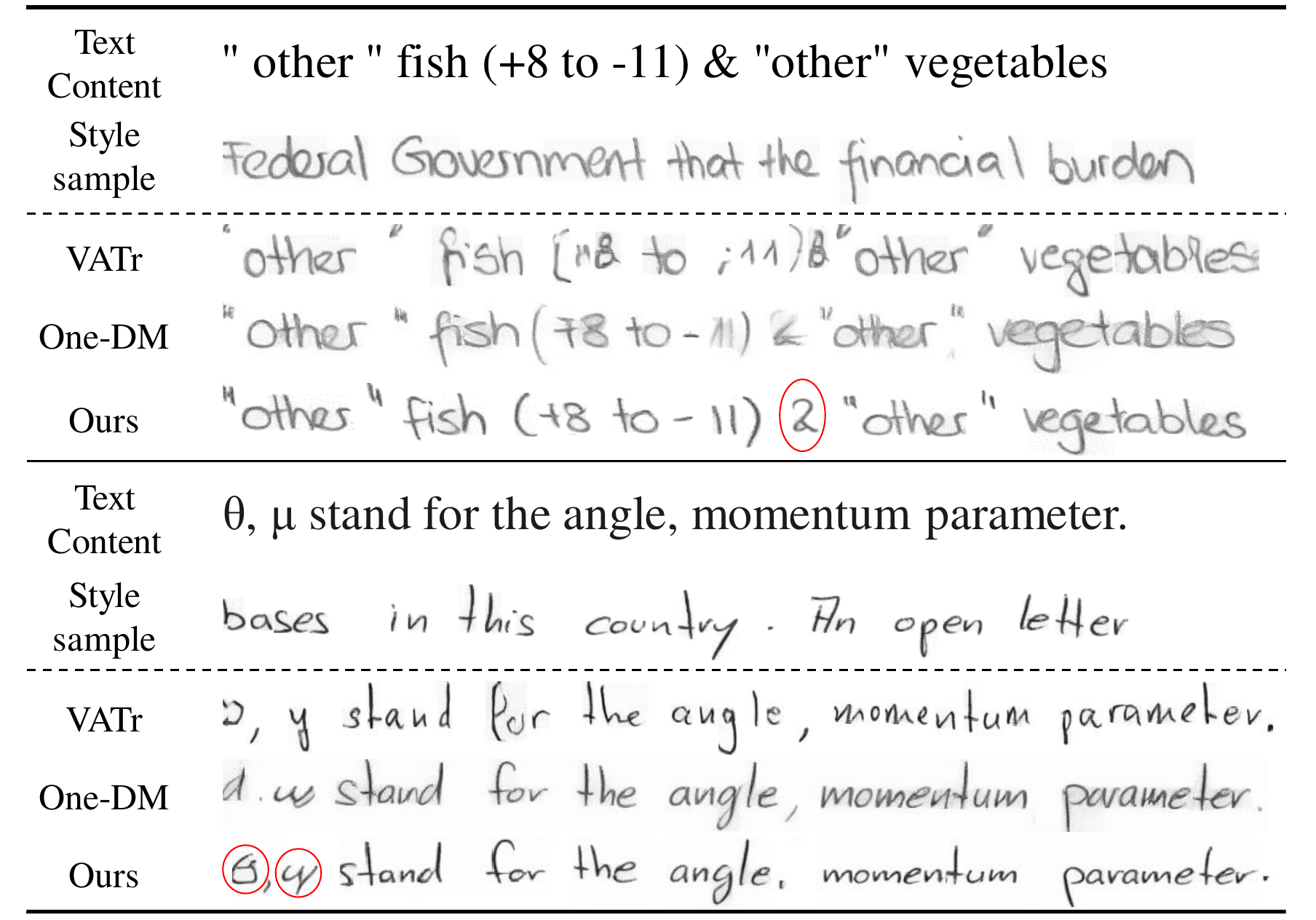}
\end{center}
\caption{Failure cases. The red circles highlight character structure errors. Better zoom in $200\%$.}
\label{fig:failure}
\end{figure}

\begin{figure}[t]  
\begin{center}
\includegraphics[width=0.8\linewidth]{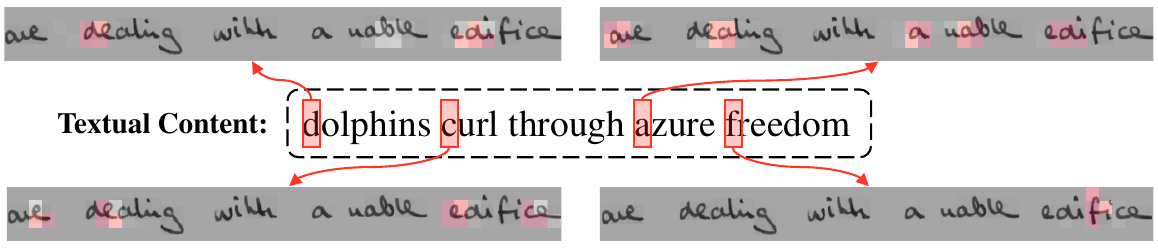}
\end{center}
\caption{Visualization results of attention maps.} 
\label{fig:attention}
\end{figure}

\section{Discussions on fine-grained style learning} 
\label{fine-grained}
Our method effectively models them for the following reasons: 1) Our content-masking strategy preserves key fine-grained features, including character-level details (cf. green circles in~\cref{fig:masking}~(c)) and stroke-level patterns (cf. purple circles in~\cref{fig:masking}~(d)). 2) Following prior study~[\textcolor{cyan}{6}], our model uses character-level content as queries in a cross-attention mechanism, enabling the style-content blender to adaptively attend to fine-grained style cues within the reference images, as shown in~\cref{fig:attention}.

\section{Analysis of failure cases}
\label{failure}
We find that DiffBrush occasionally generates structurally incorrect characters when low-frequency characters from the training set are used as content conditions. This includes punctuation marks and Greek letters, as highlighted by the red circles in~\cref{fig:failure}. A simple yet effective solution is to employ a data oversampling strategy, increasing the frequency of these characters during training.

\begin{figure}[t]
\begin{center}
\includegraphics[width=0.95\linewidth]{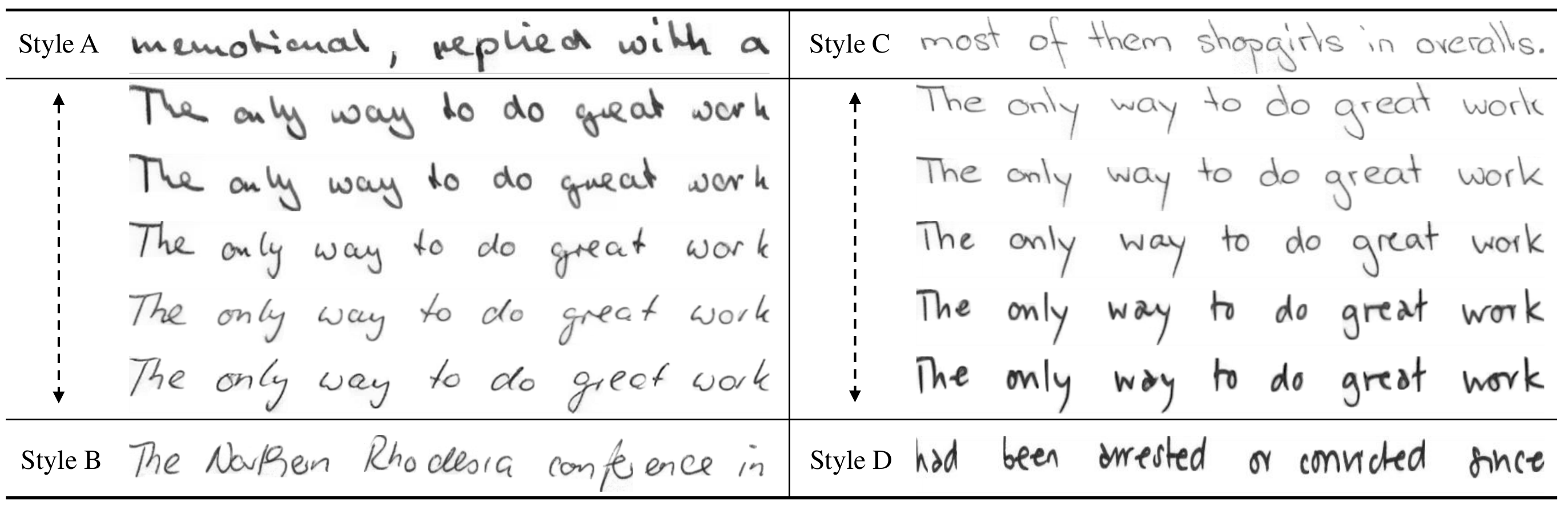}
\end{center}
\vspace{-0.15in}
\caption{Style interpolation results between different
individual handwriting styles on IAM dataset. The results are computed over both style features $S_{ver}$ and $S_{hor}$. Better zoom in $200\%$.}
\label{fig:style_interpolation}
\end{figure}

\section{Style interpolation}
\label{interpolation}
To further explore the latent space learned by our style module, we conduct linear style interpolation experiments between different writers and display the generated handwritten text-line images in~\cref{fig:style_interpolation}. From these visual results, we find that the generated text-line images smoothly transition from one style to another, in terms of character slant, and stroke thickness, while strictly preserving their original textual content. These results further confirm that our DiffBrush successfully generalizes to a meaningful style latent space, rather than simply memorizing style patterns from individual handwriting samples.

\section{More generation results}
\label{English}
\begin{figure*}[t]
\begin{center}
\includegraphics[width=0.8\linewidth]{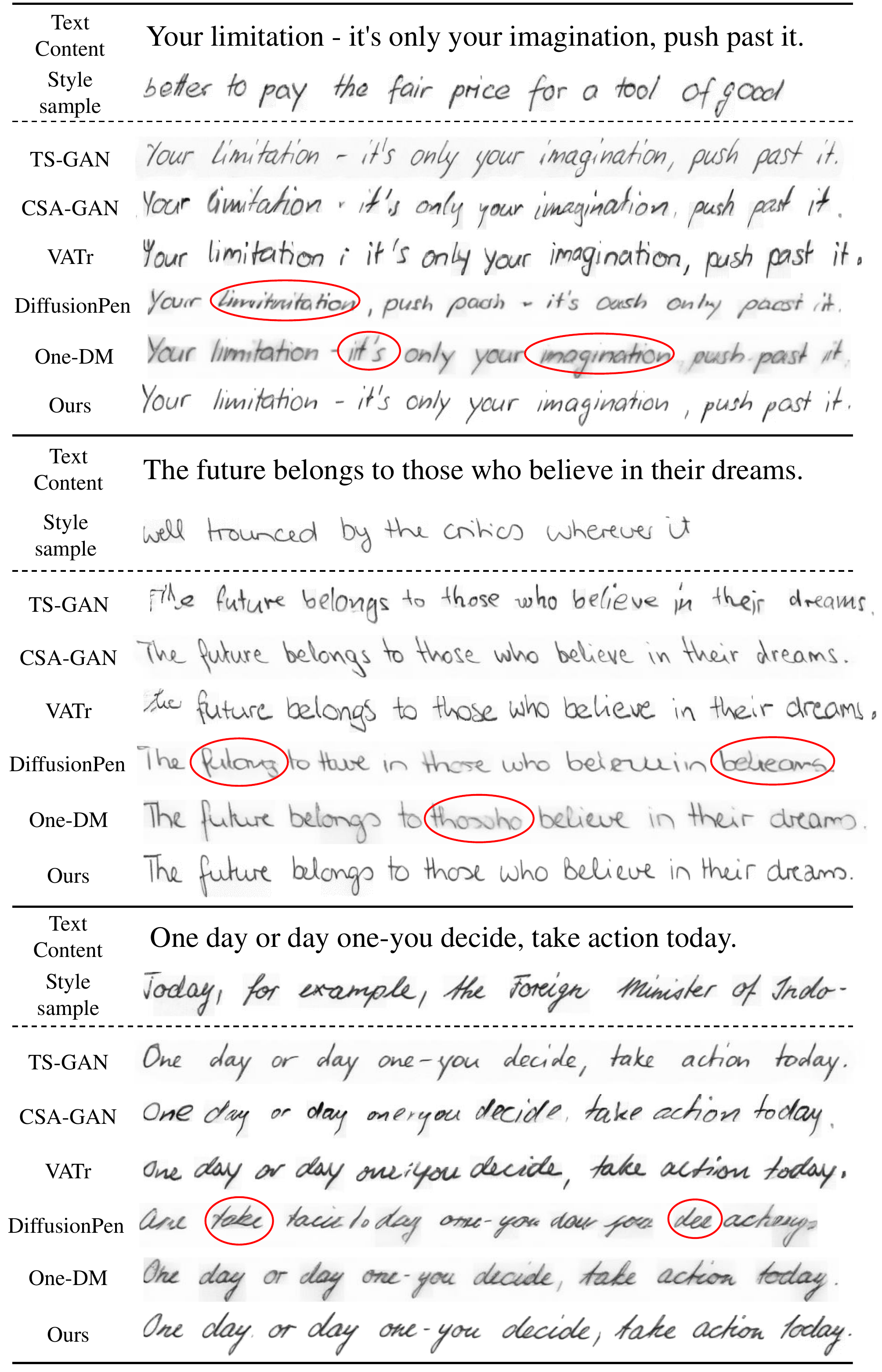}
\end{center}
\vspace{-0.1in}
\caption{Comparisons with the state-of-the-art methods for English handwritten text-line generation. The red circles emphasize incorrect content structure.}
\label{fig:English_2}
\end{figure*}

\begin{figure*}[t]
\begin{center}
\includegraphics[width=0.8\linewidth]{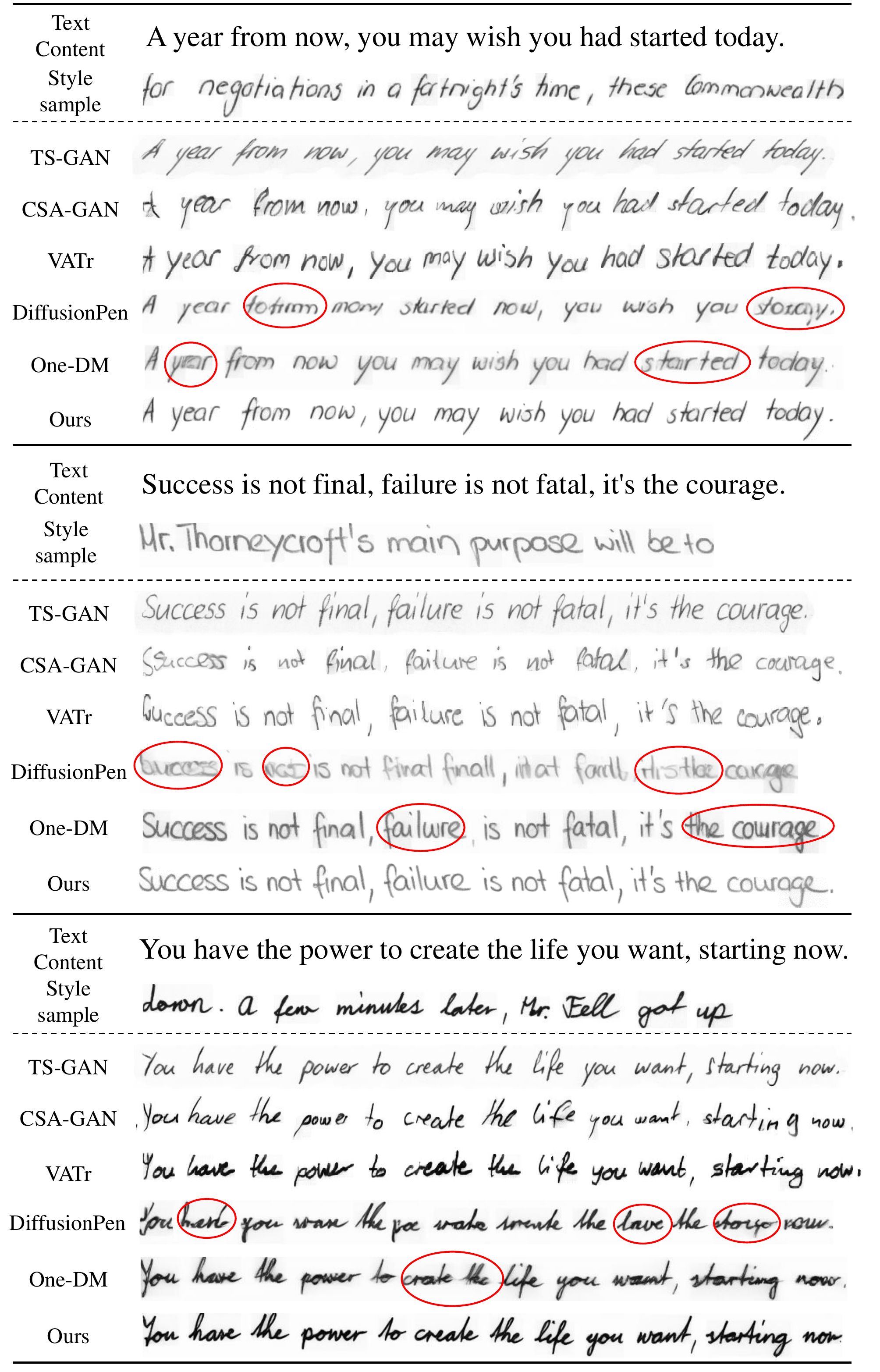}
\end{center}
\vspace{-0.1in}
\caption{Comparisons with state-of-the-art methods for English handwritten text-line generation. The red circles highlight incorrect content structure.}
\label{fig:English_3}
\end{figure*}

\begin{figure*}[t]
\begin{center}
\includegraphics[width=0.85\linewidth]{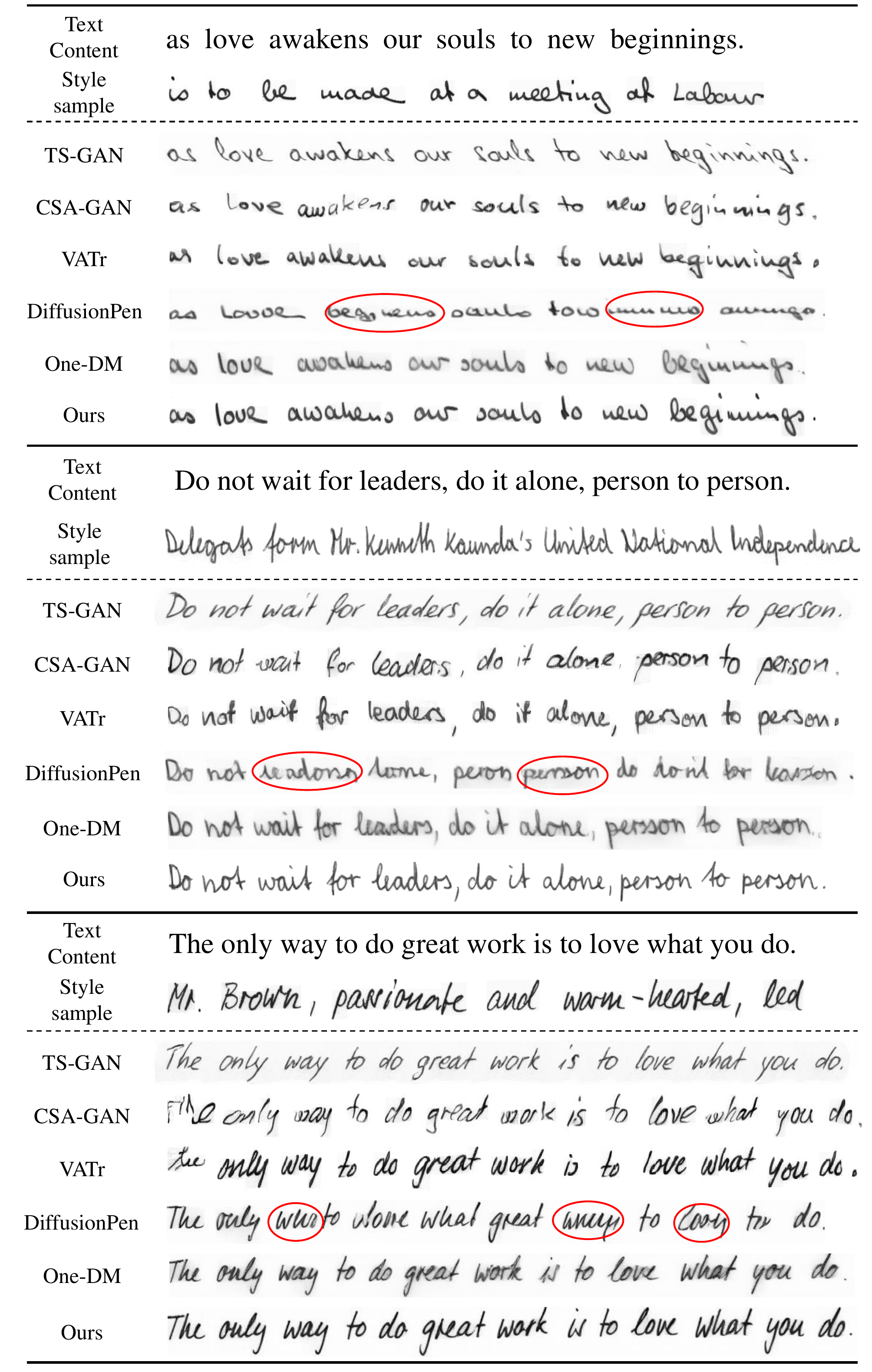}
\end{center}
\vspace{-0.1in}
\caption{Comparisons with the state-of-the-art methods for English handwritten text-line generation. The red circles highlight incorrect content structure.}
\label{fig:English_1}
\end{figure*}

\begin{figure*}[t]
\begin{center}
\includegraphics[width=0.85\linewidth]{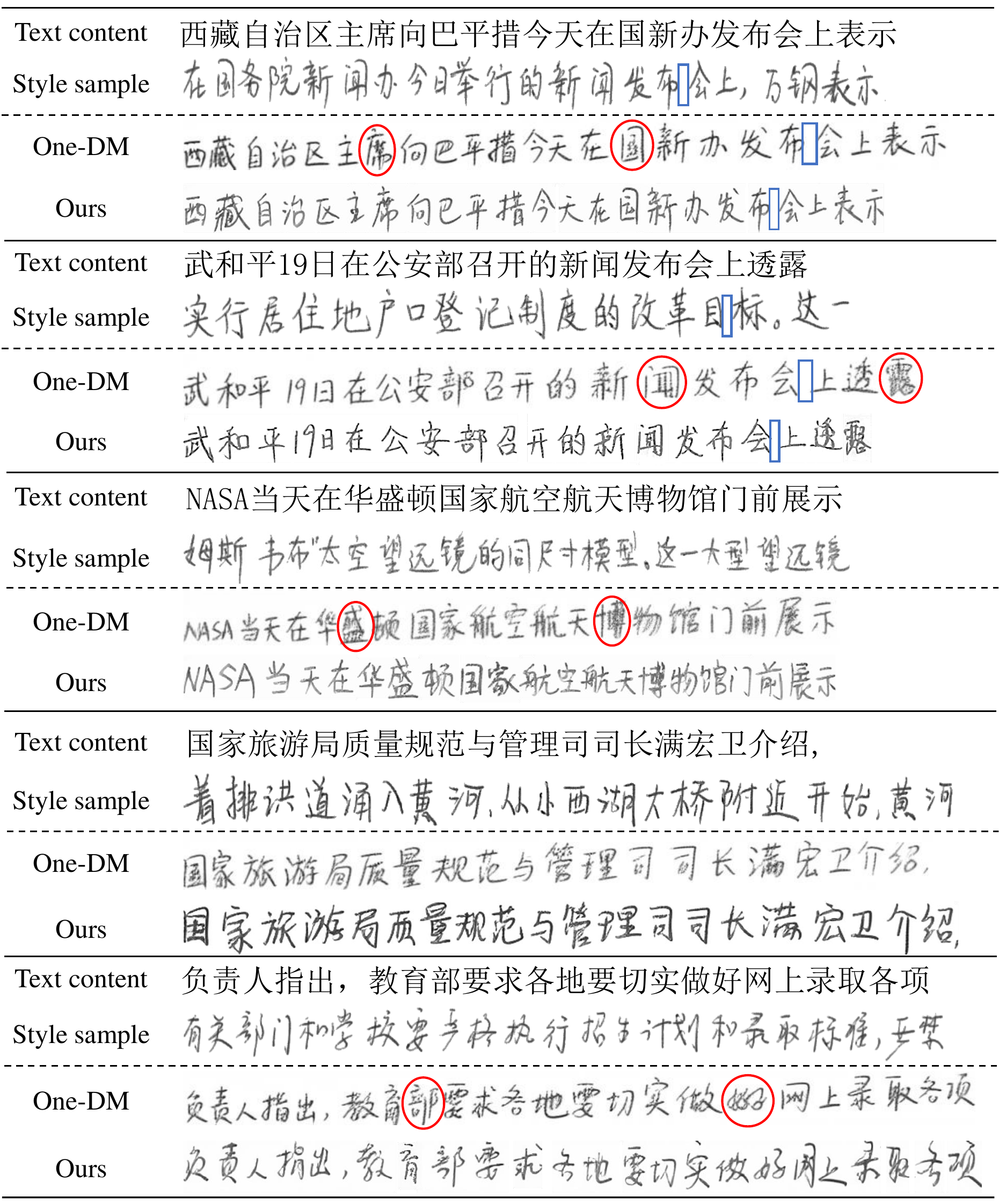}
\end{center}
\vspace{-0.1in}
\caption{Comparisons with One-DM~\cite{one-dm2024} on Chinese handwritten text-line generation. The blue boxes highlight the character spacing, while the red circles emphasize the incorrect character structures.}
\label{fig:Chinese_1}
\end{figure*}

\begin{figure*}[t]
\begin{center}
\includegraphics[width=0.85\linewidth]{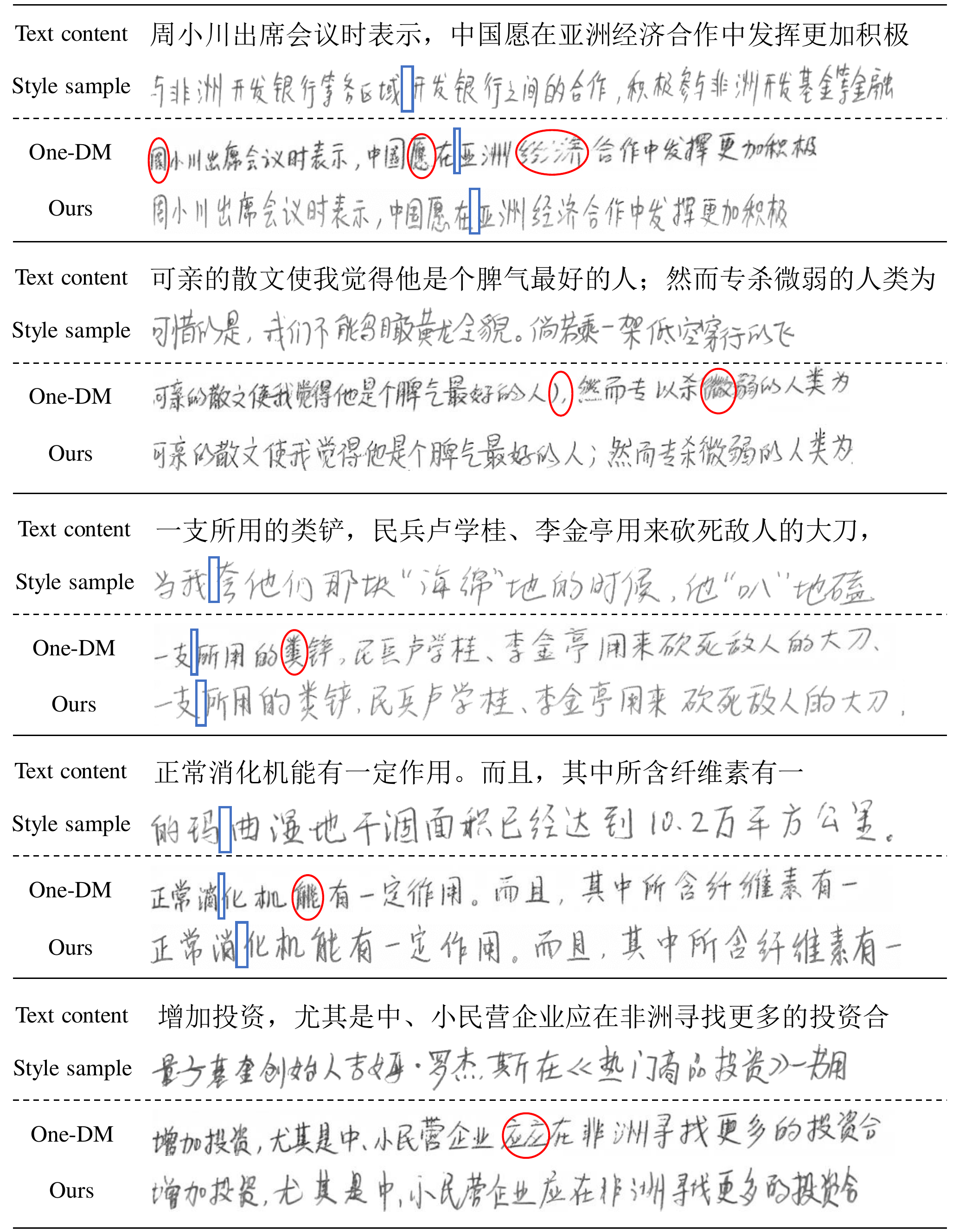}
\end{center}
\vspace{-0.1in}
\caption{Comparisons with One-DM~\cite{one-dm2024} on Chinese handwritten text-line generation. The blue boxes highlight the character spacing, while the red circles emphasize the incorrect character structures.}
\label{fig:Chinese_2}
\end{figure*}

\cref{fig:English_2}-~\cref{fig:Chinese_2} present qualitative comparisons between our DiffBrush and previous state-of-the-art methods for multilingual handwritten text-line generation, covering both English and Chinese. The extensive visual results demonstrate that our DiffuBrush excels in both style imitation and structural preservation of generated multilingual text-lines, highlighting its superior performance.

\end{document}